\documentclass{article}

\usepackage{PRIMEarxiv}

\usepackage[utf8]{inputenc} % allow utf-8 input
\usepackage[T1]{fontenc}    % use 8-bit T1 fonts
\usepackage{hyperref}       % hyperlinks
\usepackage{url}            % simple URL typesetting
\usepackage{booktabs}       % professional-quality tables
\usepackage{amsfonts}       % blackboard math symbols
\usepackage{nicefrac}       % compact symbols for 1/2, etc.
\usepackage{microtype}      % microtypography
\usepackage{lipsum}
\usepackage{graphicx}
\graphicspath{{media/}}     % organize your images and other figures under media/ folder

\usepackage{algorithm}
\usepackage{algorithmic}
\renewcommand{\algorithmicrequire}{\textbf{Initialize:}}   
\renewcommand{\algorithmicensure}{\textbf{Iteration:}} 
\renewcommand{\algorithmicinput}{\textbf{Input:}} 
\renewcommand{\algorithmicoutput}{\textbf{Output:}}

\usepackage{amsmath}
\usepackage{amssymb}
\usepackage{mathtools}
\usepackage{amsthm}

\usepackage[capitalize,noabbrev]{cleveref}
\usepackage{multirow}
\usepackage{booktabs}

\def\ie{\mbox{\textit{i.e.}, }}
\def\eg{\mbox{\textit{e.g.}, }}

\def\diam{\mbox{diam}}

\def\mA{{\mathcal A}}
\def\mB{{\mathcal B}}

\def\mD{{\mathcal D}}
\def\mE{{\mathcal E}}

\def\mL{{\mathcal L}}

\def\mN{{\mathcal N}}

\def\mR{{\mathcal R}}

\def\mU{{\mathcal U}}

\def\mW{{\mathcal W}}
\def\mX{{\mathcal X}}
\def\mY{{\mathcal Y}}

\DeclareMathAlphabet\mathbfcal{OMS}{cmsy}{b}{n}
\def\0{{\bf 0}}
\def\1{{\bf 1}}

\def\bI{{\bf I}}

\def\bu{{\bf u}}
\def\bv{{\bf v}}

\def\bx{{\bf x}}
\def\by{{\bf y}}
\def\bz{{\bf z}}

\def\mmE{{\mathbb E}}

\def\mmR{{\mathbb R}}

\def\st{{\mathrm{s.t.}}}

\def\bx{{\bf x}}

\def\by{{\bf y}}

\def\bz{{\bf z}}

\def\st{{\mathrm{s.t.}}}

\usepackage{enumitem}

\newtheorem{deftn}{Definition}
\newtheorem{thm}{Theorem}

\newtheorem{lemma}{Lemma}
\newtheorem{remark}{Remark}

\usepackage[textsize=tiny]{todonotes}

\title{Internal Wasserstein Distance \\
for Adversarial Attack and Defense
}

\author{
  Qicheng Wang\\
  Shenzhen Youjia Innov Tech Co., Ltd\\
  \texttt{wangqicheng@minieye.cc} \\
  \And  
  Shuhai Zhang \\
  South China University of Technology  \\
  \texttt{ mszhangshuhai@mail.scut.edu.cn} \\
  \And
  Jiezhang Cao \\
  South China University of Technology  \\
  \texttt{  caojiezhang@gmail.com} \\
  \And
  Jincheng Li  \\
  South China University of Technology  \\
  \texttt{   is.lijincheng@gmail.com} \\
  \And
  Mingkui Tan\footnotemark[2] \\
  South China University of Technology  \\
  \texttt{ mingkuitan@scut.edu.cn} \\
  \And
  Yang Xiang\footnotemark[2] \\
  Hong Kong University of Science and Technology \\
  \texttt{maxiang@ust.hk}\\
}

\begin{document}
\maketitle

\renewcommand{\thefootnote}{\fnsymbol{footnote}}
\footnotetext[2]{Corresponding authors.}
\renewcommand{\thefootnote}{\arabic{footnote}}

\begin{abstract}
  Deep neural networks (DNNs) are known to be vulnerable to adversarial attacks that would trigger misclassification of DNNs but may be imperceptible to human perception. Adversarial defense has been an important way to improve the robustness of DNNs. Existing attack methods often construct adversarial examples relying on some metrics like the $\ell_p$ distance   to perturb samples. However, these metrics can be insufficient to conduct adversarial attacks due to their limited perturbations. In this paper, we propose a new internal Wasserstein distance (IWD) to {capture the semantic similarity of two samples}, and thus it helps to obtain larger perturbations than 
  {currently used metrics such as the $\ell_p$ distance}.
  {We then apply the internal Wasserstein distance to perform adversarial attack and defense.} In particular, we develop a novel attack method relying on IWD to {calculate the similarities between an image and its adversarial examples.
  } In this way, we can generate { diverse and semantically similar} adversarial examples that are more difficult to defend by existing defense methods. Moreover, we devise a new defense method relying on IWD to learn robust models against unseen adversarial examples. We provide both thorough theoretical and empirical evidence to support our methods.
\end{abstract}

\section{Introduction} \label{sec:introduction}
{Deep neural networks (DNNs) are vulnerable to adversarial examples \cite{szegedy2013intriguing,pal2020game,Croce2021MindTB,Pintor2021FastMA} mainly due to their overfitting nature \cite{rice2020overfitting}.
Particularly, this issue becomes more severe if we have only limited training data for a specific task.
In fact, the neural network tends to fit the training data well and may make an incorrect prediction {for} an example contaminated by only slight perturbations or transformations. {Moreover}, {the} security of deep learning systems {is} vulnerable to crafted adversarial examples, which may be imperceptible to the human perception but can lead the model to misclassify the output~\cite{papernot2017practical,  sharif2016accessorize}.} 
In practice, adversarial examples can be generated by perturbing or transforming some image pixel values to ensure they {maintain the semantic similarity with}
the original images. Such similarity
is a perceptual metric {that} is {often defined by a simple metric such as the $\ell_p$ distance \cite{goodfellow2014explaining}.}
{However, this  $\ell_p$ metric 
suffers from two limitations.}

	\begin{figure}[t]
		\centering
		%		\vspace{-20pt}
		\includegraphics[width=0.5\textwidth]{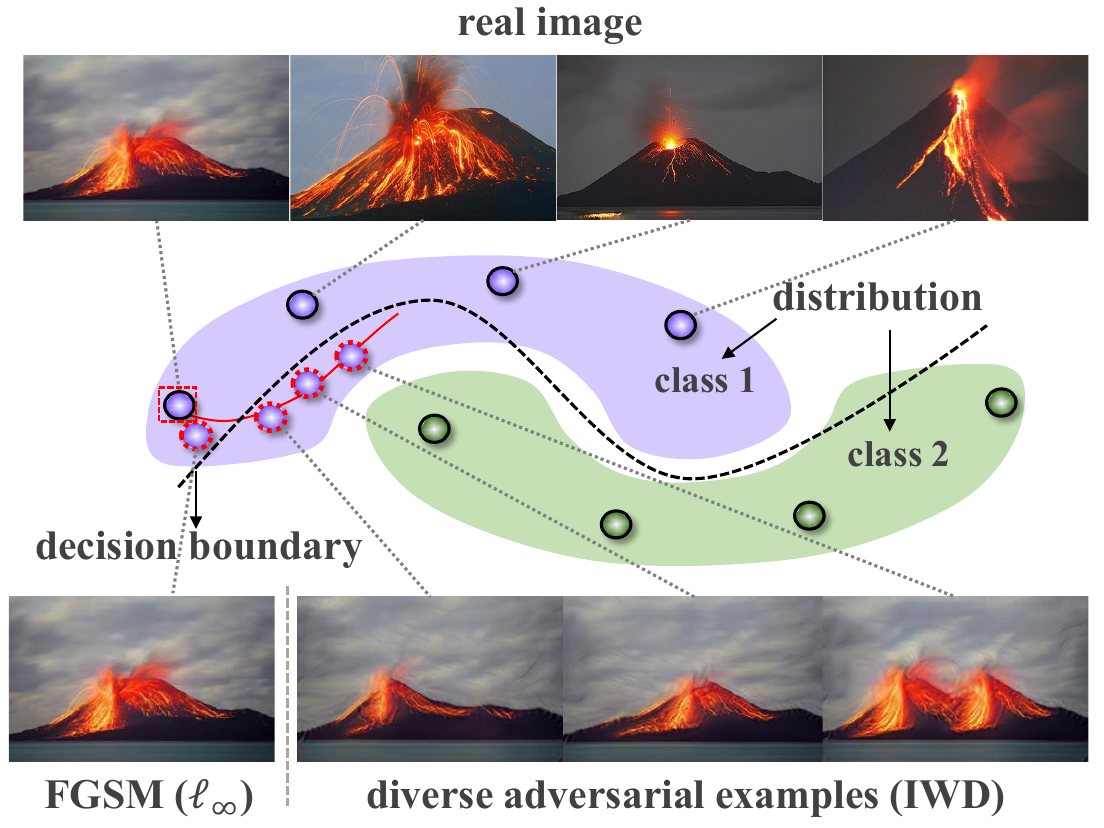}
		\vspace{-2pt}
		\caption{
		Adversarial examples generated with $\ell_\infty$ and IWD using FGSM and our method. 
		{
		The $\ell_\infty$ adversarial examples are close to the original examples, may result in an unsuccessful attack when the sample is far from the decision boundary. In contrast, the IWD adversarial examples are far away from the original ones while keeping the semantic similarity of the samples.
		}
		}
		\label{fig:example_diverse_AE}
		\vspace{-7pt}
	\end{figure}

First, most attack methods \cite{goodfellow2014explaining, madry2017towards, Croce2021MindTB} use the $ \ell_p $ distance as a similarity metric to fool a classifier. However, the $ \ell_p $ metric, despite its computational convenience, is  ineffective to elaborate all possible adversarial perturbations (\eg translation, dilation or deformation) due to its insufficient perturbations.
Specifically, the attackers always use this metric to ensure that adversarial examples are bounded in a small norm box to keep the semantic consistency of adversarial examples and original ones, which limits the effectiveness of the adversarial perturbations. 
{As shown in Figure~\ref{fig:example_diverse_AE}, adversarial examples crafted with $\ell_\infty$ using FGSM \cite{goodfellow2014explaining} are often close to the original sample, which may result in an ineffective attack when the sample is distant from the decision boundary.}
Thus, it is difficult for each image to find an effective adversarial example that causes misclassification of the classifier. 
Therefore, how to measure the similarity of a sample and its adversarial example is very necessary and important.

Second, models trained merely on $\ell_p$ adversarial examples are vulnerable to the adversarial examples crafted on the manifold with generative models \cite{Song2018ConstructingUA,Poursaeed2019FinegrainedSO}. One possible reason is that those attack methods, used for generating $ \ell_p $ adversarial examples, understand an input image from a one-sided view in Euclidean space and thus may be insufficient to perturb samples on the manifold. In this sense, 
perturbing a sample on the manifold would  be important to achieve a robust model. In this way, we would obtain more diverse and valuable adversarial examples for training. 
{More importantly, diverse adversarial examples} can help in covering the blind area of the data distribution to develop robust models \cite{Croce2020ReliableEO,Madaan2021LearningTG}.
Unfortunately, how to generate diverse adversarial examples and build robust models with them is unknown.

{In this paper, we investigate two key questions: 1) how to effectively attack a model on the manifold; {and} 2) how to build robust models relying on those attacks. Nevertheless, we find the commonly used $\ell_p$ distance is insufficient when conducting attack and defense to DNNs on the manifold. To address this, we propose an internal Wasserstein distance (IWD). Relying on IWD, we develop a new attack method (called IWDA) to perturb samples on the manifold.	Correspondingly, we develop a defense method (called IWDD) to defend against unseen adversarial examples.}

The contributions of this paper are summarized as follows:
\begin{itemize}[leftmargin=*]
	\item We analyze the reasons why the $\ell_p$ distance is insufficient when doing attacks and defenses.
	We thus propose an internal Wasserstein distance (IWD) to exploit the internal distribution of data for both attack and defense tasks.
	\item We propose a novel attack method to craft  
	adversarial examples using the proposed IWD metric.
	Quantitative and qualitative experiments demonstrate the superiority of the proposed method.
	\item We theoretically derive an upper bound for the attack classification error, and this motivates us to devise a new defense method relying on IWD, to improve the robustness of the classifier.
\end{itemize}

\section{Related Work} \label{sec:related_work}

\subsection{Adversarial Attack}
\textbf{Attack with {the} $\ell_p$ perturbation.}
DNNs are vulnerable to adversarial examples with a
perceptible perturbation \cite{szegedy2013intriguing}.
Most existing methods \cite{goodfellow2014explaining,Croce2020ReliableEO,croce2020minimally,athalye2018synthesizing,Croce2021MindTB} attempt to craft adversarial examples with the $\ell_p$ perturbation.
However, these adversarial examples cannot result in a large perturbation that can guarantee successful attacks.
Besides, the $\ell_p$ perturbation is not a good metric of image similarity \cite{alaifari2018adef}.

\textbf{Attack with the {non-}$\ell_p$ perturbation.}
To mitigate the limitations of {the} $\ell_p$ perturbations, several methods \cite{alaifari2018adef,liu2018beyond,athalye2018synthesizing,zhao2018generating,xiao2018spatially,icmlWuWY20} try to craft adversarial examples {beyond the $\ell_p$ perturbations.} 
To this end,
\cite{alaifari2018adef} propose an ADef algorithm to craft adversarial examples by deforming images iteratively.
{In addition, \cite{wong2019wasserstein} iteratively
project adversarial examples onto the Wasserstein ball. To develop stronger and faster attacks, \cite{icmlWuWY20} apply projected gradient descent and Frank-Wolfe to craft adversarial examples.}
However, these adversarial examples may lack diversity. In this paper, we propose a novel metric, internal Wasserstein distance (IWD), to generate diverse adversarial examples.

\subsection{Adversarial Defense}
To address the vulnerability of DNNs, many defense methods \cite{goodfellow2014explaining, kannan2018adversarial} have been proposed to defend against adversarial examples. For example, 
\cite{madry2017towards} present 
a projected gradient descent (PGD) training method to improve the resistance of the model to a wide range of attacks. 
Unfortunately, training with PGD is time-consuming and computationally intensive \cite{shafahi2019adversarial}. 
To address this, free-PGD \cite{shafahi2019adversarial} and fast-AT \cite{Wong2020FastIB} speed up the training and improve the robustness of the model.

Recently, PGD adversarial training is a popular method to defend against adversarial examples. However, most studies \cite{zhang2019theoretically,Raghunathan2020,Yang2020,zhang2020fat,wang2019improving,Madaan2021LearningTG} posit that a robustness-accuracy {trade-off} may be inevitable in defense methods. For example, \cite{Mehrabi2021FundamentalTI} theoretically analyze the trade-off between standard risk and adversarial risk and derive a Pareto-optimal trade-off over some specific classes of models in the infinite data with fixed features dimension. 
In addition, \cite{Sriramanan2021TowardsEA} introduce a Nuclear-Norm regularizer to enforce the function smoothing in the vicinity of data samples to achieve an efficient and effective adversarial training method.
Moreover, \cite{Robey2021AdversarialRW} propose a hybrid Langevin Monte
Carlo technique to improve the performance of adversarial training.
However, these methods are limited {in improving} the robustness of a classifier due to the limitations of the $\ell_p$ distance. In this paper,
we derive an upper bound to enhance the robustness relying on the IWD.

\section{Notation and Motivations} \label{sec:problem_definition}
\begin{figure}[h]
    \centering
    \vspace{-2pt}
    \includegraphics[width=0.44\textwidth]{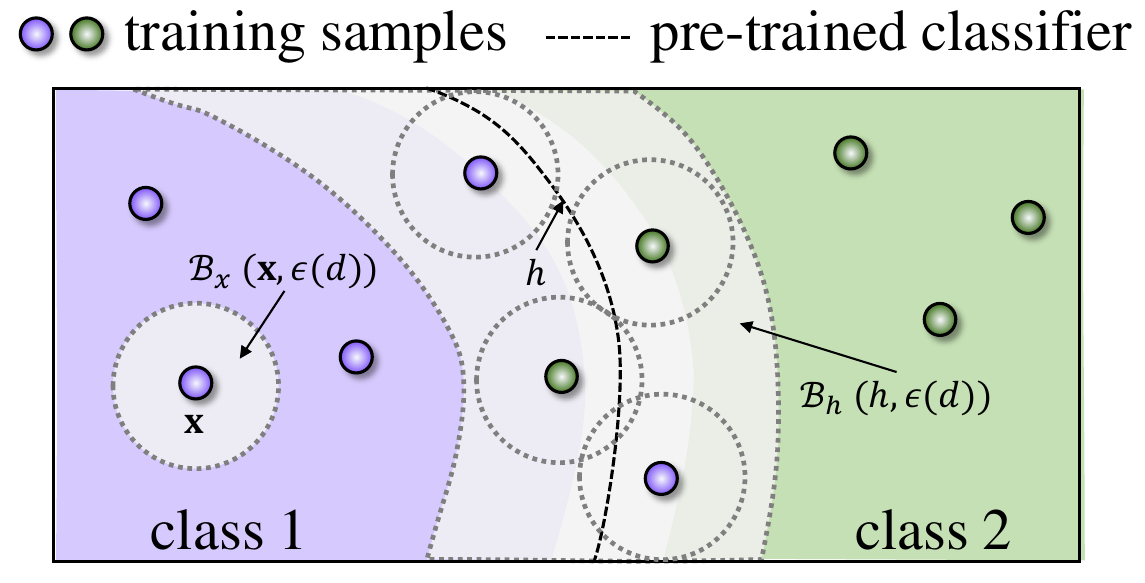}
    \vspace{-2pt}
    \caption{Illustration of notations.}
    \label{fig:notation}
    \vspace{-5pt}
\end{figure}

\paragraph{Notation.}
We use calligraphic letters (\textit{e.g.}, $ \mX $) to denote spaces, and bold lower case letters (\textit{e.g.}, $ \bx $) to denote vectors.
{Without loss of generality, we first consider a framework of binary classification, which can be well generalized to the multi-class classification.}
Let $\mD$ be the real data distribution, and
$ {\widehat{\mD}} {=} \{(\bf x_i, \bf y_i)\}_{i=1}^n {\in} \mX {\times}  \mY $ be the training data, 
{where $\bx \small{\in} \mX \small{\subset} \mmR^m$ denotes the data and $\by \small{\in} \{-1,+1\}$ denotes the label.}
Let $ h: \mX {\to} \mY $ be a classifier learned on $\widehat{\mD}$ {and valued in $\{-1, +1\}$}. Let $\diam(\mA) $ be the diameter of a set $\mA$.
Let $ \1{\{A\}} $ be an indicator function, where $ \1{\{A\}}\small{=}1 $ if $A$ is true and $ \1{\{A\}}\small{=}0 $ otherwise.
Let $\epsilon(d)$ be a sufficiently small constant with respect to some metric $d$, and $\mB_x(\bx, \epsilon(d)) \small{=} \{ \widetilde{\bx} \in \mX {\mid} d(\bx, \widetilde{\bx}) \leq \epsilon(d)\}$ 
be the $\epsilon(d)$-neighborhood set of $\bx$. 
Similarly, let $ \mB_h(h, \epsilon(d)) {=} \{ \bx {\in} \mX {\mid}\exists\; \widetilde{\bx} {\in} \mB_x(\bx, \epsilon(d)) \;\st\; h(\bx) {\neq} h(\widetilde{\bx}) \} $ be the $\epsilon(d)$-neighborhood set of $h$, as shown inFigure \ref{fig:notation}.

To develop our proposed method, we first provide the definition of the adversarial example as follows.

\begin{deftn} \emph{\textbf{($ \epsilon(d)$-adversarial example)}}	\label{def:adversarial_example}
	Given a sample pair $ (\bx, \by) $,
	the sample $ \bx $ admits an  $\epsilon(d)$-adversarial example if there exists a sample
	$ \widetilde{\bx} {\in} \mX $ such that $ h(\widetilde{\bx}) {\neq} \by $ and $ d(\bx, \widetilde{\bx}) {\leq} \epsilon(d), \epsilon(d){>}0 $.
\end{deftn}

To construct an adversarial example $\widetilde{\bx}$, most existing methods often
use the $\ell_p$ distance as the metric $d$. 
However, this metric is insufficient when conducting attack and defense.

\textbf{Limitations of the $\ell_p$ distance {for attack and defense}.}
The $\ell_p$ distance may only 
ensure that the crafted adversarial examples are close to the original sample due to small perturbations (seeFigure \ref{fig:example_diverse_AE}). 
When performing the attack, if a sample is far away from the decision boundary, 
it is difficult
to find an effective adversarial example that can 
fool the classifier. Thus, the $\ell_p$ distance is insufficient to conduct adversarial attacks when a good classifier can well describe the boundaries of different classes with a large margin.

For defense, most existing methods reinforce the training procedure by searching for adversarial examples relying on the $\ell_p$ distance. In this way, they try to learn a robust classifier to cover the blind area of the data distribution.  
However, they may fail to cover the whole data distribution due to {the} limited range of {the} $\ell_p$ perturbation, leading to inferior defense performance. {To address these, we propose a new metric in this paper to exploit the internal distribution of data for both attack and defense.}

\subsection{Internal Wasserstein Distance}\label{section3.1}
As aforementioned, the attack and defense methods relying on 
 the $\ell_p$ distance are insufficient to perturb the samples.
 Recently, the Wasserstein distance \cite{villani2008optimal} has been proposed to measure the differences between two given distributions and widely used in generated adversarial networks (GANs) \cite{WGAN}.

\begin{deftn} \emph{\textbf{(1-Wasserstein distance \cite{villani2008optimal})}} 
	Given two distributions $ \mD $ and $ \mD' $ of $\bx$ and $\bx'$, respectively, 
	the Wasserstein distance can be defined as:
	\begin{align}
	\widetilde{\mW}\left(\mD, \mD' \right)=\inf _{\mathrm{T} \in \Pi\left(\mD, \mD'\right)} \mathbb{E}_{(\bx, \bx') \sim \mathrm{T}}[\|\bx- \bx'\|_1].
	\label{def:WD}
	\end{align}
	Here, $\Pi$ is a set of all joint distributions $\mathrm{T}$ whose marginal distributions are $\mD$ and $\mD'$.
\end{deftn}

Here, {we consider %p{=}1$ in
	$1$-Wasserstein distance} following \cite{WGAN}.
	Concretely, the 1-Wasserstein distance has a well-defined dual formulation  and can be developed in WGAN \cite{WGAN} using the Kantorovich-Rubinstein duality \cite{villani2008optimal}.
{Note that  while the Wasserstein distance has been widely used in GANs, it is non-trivial to  apply it  to conduct adversarial attack, since the Wasserstein distance between two samples in Eqn.~(\ref{def:WD}) is difficult to be defined without defining their distributions. To this end, some methods \cite{wong2019wasserstein,icmlWuWY20} model it as the moving pixel mass from one whole image to another with EMD distance, which, however, may 
fail to measure the similarity of images when the internal patch of the image transforms, \eg swapping the internal patch.}

To address this, we propose an internal Wasserstein distance (IWD) to measure the internal distribution distance between a sample $\bx$ and its adversarial sample $\widetilde{\bx}$. 	Specifically, let $\{\bu_i\}_{i=1}^N$ and $\{\bv_i\}_{i=1}^N$ be $N$ patches drawn from two samples $\bx$ and $\widetilde{\bx}$, respectively.  
We define the internal distributions  of $\bx$ and $\widetilde{\bx}$ as $\mu{=}\frac{1}{N} \sum_i \delta_{\bu_i}(\bx)$ and $\nu{=}\frac{1}{N} \sum_i \delta_{\bv_i}(\widetilde{\bx})$, respectively, where $\delta$ is the Dirac distribution~\cite{alt2006lineare}. Here, we select patches from different scaled images. Specifically, for an image $H{\times}W (H\leq W)$, let $H_{min}$ be the minimum height of the scaled image and $P_{min}$ be the minimum size of the patches. Given  a scale set $\mathcal{S}{=}\{j \mid s^j{\leq}H_{min}/H, j{\in} \mathbb{N}\}$ with a scale factor $s$, we decompose each scaled image $(H/j{\times}W/j)$ uniformly into $2^\gamma {\times} 2^\gamma$ patches, where $\gamma \in \{\gamma \mid (H/j)/{2^\gamma} \geq P_{min}\}$ is the sampling rate.
Then, we define the IWD as follows.

\begin{deftn} \emph{\textbf{(Internal Wasserstein distance)}} \label{def:def_IWD}
	Given internal distributions $ \mu $ and $ \nu $ of $ \bx $ and $ \widetilde{\bx} $, respectively, the internal Wasserstein distance can be defined as:
	\begin{align}
	\label{def:IWD}
	\mW(\bx, \widetilde{\bx}) := \inf\nolimits_{\gamma \in \Gamma(\mu, \nu)} \mmE_{(\bu, \bv) \sim \gamma} \left[ \| \bu - \bv \|_1 \right],
	\end{align}
	where $ \Gamma $ is a set of all joint distributions $ \gamma$ whose marginal distributions are $ \mu $ and $ \nu $.
\end{deftn}

\begin{remark}
	\label{remark_WD_IWD}
	{From the definition, 
	IWD is different from the Wasserstein distance. Specifically, the Wasserstein distance in Eqn.~(\ref{def:WD}) is not able to directly calculate the similarity between two samples $\bx$ and $\widetilde{\bx}$ without defining their distributions. Although \cite{wong2019wasserstein} and \cite{icmlWuWY20} consider this problem as the moving mass form  $\bx$ to $\widetilde{\bx}$, this  is not necessarily a good measure of image similarity: \eg swapping the internal patches of $\bx \small{=} [\bx_1, \bx_2] $ causes their Wasserstein distance
	$\widetilde{\mW} \left([\bx_1, \bx_2], [\bx_2, \bx_1]\right)	\small{>} 0$ if  $\bx_1 \small{\neq} \bx_2$, which will be zero in our IWD, \ie ${\mW} \left([\bx_1, \bx_2], [\bx_2, \bx_1]\right)	\small{=} 0$.}
\end{remark}

Definition \ref{def:def_IWD} indicates that IWD {captures the semantic similarity of two samples with the help of defining distributions of patches in the image}, 
which thus is able to perturb a sample on the manifold when 
performing attacks.
The following theorem indicates that the IWD can lead to larger transformations than the pixel perturbation (\eg the $\ell_p$ distance).

\begin{thm} \label{thm:ball_W_lp}
    {
	Let $\epsilon(\ell_p)$ and $\epsilon(\mW)$ denote the perturbation constants with the $\ell_p$ distance and IWD, respectivly.
	Then for any $\epsilon(\ell_p) \small{>} 0 (p \small{>} 0)$, there exists $\epsilon(\mW) \small{\leq} \epsilon(\ell_p)$ satisfies:}
	\begin{align*}
	\diam (\mB_x(\bx, \epsilon(\ell_p))) \leq \diam (\mB_x(\bx, \epsilon(\mW))).
	\end{align*}
\end{thm}

{Theorem \ref{thm:ball_W_lp} shows that the diameter of the perturbation ball in terms of $\epsilon(\mW)$ is larger than that in terms of $\epsilon(\ell_p)$.}
Thus, IWD would have a higher probability to find valid adversarial examples that fool the model than the $\ell_p$ distance. {In the following, we apply IWD to perform attack and defense.}

\section{Adversarial Attack with IWD}
\label{sec:proposed_method}

In general, the \textbf{adversarial attack} aims to learn an adversarial attacker $  g(\bx, \bz)$ to confuse a classifier $h$, where $\bz$ can be a random noise vector or other kinds of input. 
We consider untargeted attack and targeted attack,
which can be formulated {as} the following optimization problems:
\begin{equation}
\begin{aligned}
&\underbrace{\max\nolimits_{g}\;\;  \mL(h(\widetilde{\bx}) \by)}_{\text{untargeted attack}}  \quad \text{or} \quad \underbrace{\min\nolimits_g\;\;  \mL(h(\widetilde{\bx}) \by_t)}_{\text{targeted attack}},
\label{prob:adv_attack}
\end{aligned}
\end{equation}
where $\widetilde{\bx}=g(\bx, \bz) \in \mB_x(\bx, \epsilon(d))$, $\by$ is the label of $\bx$ and $\by_t$ is the label of the targeted attack,  {$\mL(\cdot)$ is some classification-calibrated loss \cite{bartlett2006convexity}.}

Relying on IWD, we are able to 
craft semantically similar but diverse adversarial examples.
For example, the crafted adversarial examples of a volcano image (see Figure~\ref{fig:example_diverse_AE}) contain one or more volcanoes with different sizes, shapes and locations.
However, it is non-trivial to directly apply IWD to perform adversarial attack to obtain such adversarial examples, since the infimum in Eqn.(\ref{def:IWD}) is highly intractable \cite{WGAN}. To conduct adversarial attack with IWD, we next introduce the  objective function for attack.

    \begin{figure}[t]
		\centering
		{
			\includegraphics[width=0.6\linewidth]{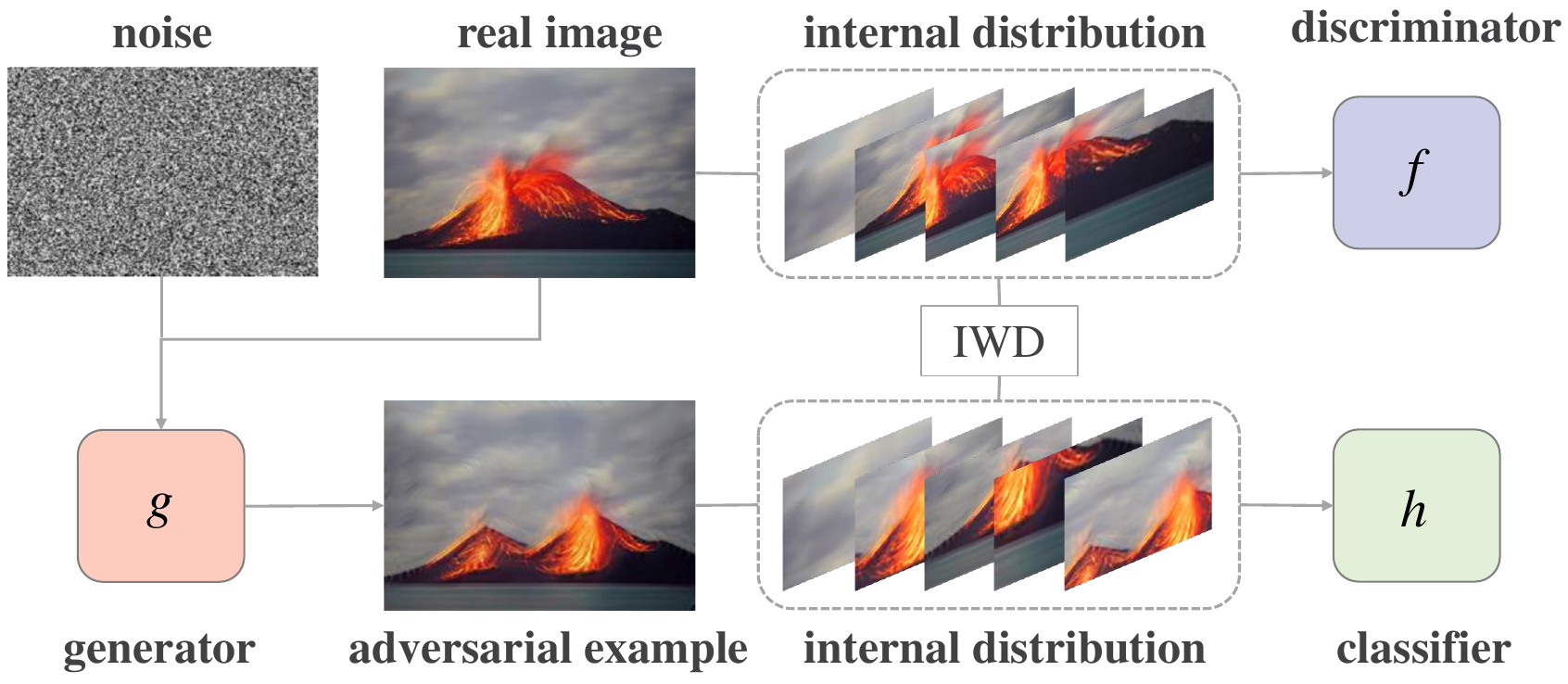}
			\vspace{-2pt}
			\caption{Illustration of our attack method (IWDA). Specifically, $g$ produces an adversarial example such that $f$ cannot distinguish the internal distribution of patches from real/fake data, 
			{leading to misclassification of the classifier $h$.}}
			\label{fig: framework}
		}
	\end{figure}
	
\subsection{Objective Function for Attack}
To generate $\epsilon(\mW)$-adversarial examples, we propose a new attack method relying on IWD, called IWDA.
Figure \ref{fig:example_diverse_AE} (a) illustrates the overall scheme of IWDA.
Specifically, we optimize an adversarial attack loss $\mL_{\text{adv}}(f, g)$ to craft $\epsilon(\mW)$-adversarial examples.
Meanwhile, we optimize a classification loss $\mL_{\text{c}} (g; h)$ such that the classifier misclassifies the adversarial examples.
Then, the total objective function can be written as follows:
\begin{align}
\mL(f, g) = \mL_{\text{adv}}(f, g) + \tau \mL_{c} (g; h), \label{problem:total}
\end{align}
where $ \tau $ is a hyper-parameter.
Next, we introduce the adversarial attack loss and classification loss. 

\textbf{Adversarial attack loss.}
In Figure \ref{fig: framework}, the generator $ g $ aims to produce an adversarial example $ \widetilde{\bx} = g(\bx, \bz) $, and then the discriminator $ f $ distinguishes the internal distributions of patches in the adversarial example $ \widetilde{\bx} $ and the original image $ \bx $.
Relying on the internal Wasserstein distance, we propose to optimize the following adversarial loss:
\begin{align}
\label{def:adversarial_loss}
\mL_{\text{adv}}(f, g) {=}\! \max_{\|f\|_{\!L}{\leq} 1} \mmE_{\bu{\sim} \mu} \!\left[ f(\bu) \right] {-} \mmE_{\bv{\sim} \nu} \!\left[ f({\bv})\right],
\end{align}
where 
$ \mu $ and $ \nu $ denote the internal distributions of patches in $ \bx $ and $ \widetilde{\bx} $, respectively.
Here, $\| f \|_L$ is the Lipschitz constant of $f$.
In practice, we introduce a gradient penalty  \cite{Gulrajani2017gangp}  $ \mR_{\text{gp}}(f) {=} \hat{\mmE} [ (\| \nabla f(\hat{\bx}) \| {-} 1)^2 ] $ to optimize Problem (\ref{def:adversarial_loss}).
Note that we do not necessarily constrain IWDA adversarial examples bounded in an IWD ball from original images. The IWD distance between adversarial examples and original ones decreases during the
training.
In Section \ref{sec:attack}, we measure the perturbation distance under the IWD metric. The last sub-figure of Figure \ref{fig:l2_distance} shows the distance between IWDA adversarial examples and the original counterparts is close to zero on average under the IWD metric.
	\begin{algorithm}[t]
	\caption{Attack method (IWDA).}
	\label{alg:attack}
	\begin{algorithmic}[1]\small
		\INPUT  A sample $\bx$, the hyper-parameters $ \lambda $ and $ \tau $, the number of iterations of the discriminator per generator iteration $ n_{\text{critic}} $. \\
		\OUTPUT The discriminator $ f $, the generator $ g $,
		\WHILE{not converged} 
		\FOR{ $ t= 0, \ldots, n_{\text{critic}} $} 
		\STATE Adversarial examples $ \widetilde{\bx} {=} g(\bx, \bz) $, where $\bz {\sim} \mN(\0, \bI) $
		\STATE Sample $ \hat{\bx} \leftarrow \rho \bx + (1-\rho) \widetilde{\bx} $, where $ \rho \sim \mU(0, 1)$  \\ 
		\STATE Gradient penalty: $ \mR_{\text{gp}} (f) = \hat{\mmE} \left[ (\| \nabla_{\hat{\bx}} f(\hat{\bx}) \|_2 {-} 1)^2 \right] $
		\STATE Update discriminator $ f $ by ascending the gradient: \\
		\vspace{5pt}
		~$\nabla_w\! \left[ \mathop{\hat{\mmE}}\nolimits_{\bu {\sim} \mu} [f(\bu)] {-} \mathop{\hat{\mmE}}\nolimits_{\bv {\sim} \nu} [f(\bv)] {+} \lambda \mR_{\text{gp}} (f) \right]$  \\
		\vspace{5pt}
		\ENDFOR
		\STATE Update the generator $ g $ by descending the gradient: \\
		\vspace{3pt}
		~~~~~~~~~~~~~~~~$\nabla_{v} \left[ - \mathop{\hat{\mmE}}\nolimits_{\bv {\sim} \nu} f({\bv}) + \tau \mL_{c} (g; h) \right]$\\
		\ENDWHILE
	\end{algorithmic}
	\end{algorithm}

\textbf{Classification loss.}
The classification loss $\mL_{\text{c}} (g; h)$ can be calculated by the distance between the prediction $h(g(\bx, \bz))$ and the ground truth $ \by $ (untargeted attack),
or the opposite of the distance between the prediction and the target class $ \by_t $ (targeted attack).
Then, $\mL_{\text{c}} (g; h)$ can be written as:
\begin{align}
\mL_{c} (g; h) = \left\{
\begin{aligned}
- \mL(h(g(\bx, \bz)) \by),   \; &\text{untargeted attack}, \\
\mL(h(g(\bx, \bz)) \by_t), \; &\text{targeted attack}.
\end{aligned}
\right.
\end{align}
The detailed algorithm is shown in Algorithm \ref{alg:attack}.

{In practice, we select different patches of images by training a pyramid GAN model with different scales progressively in a coarse-to-fine manner. For example, we first set a sliding window (e.g. kernel size = 3) and select $N$ patches of the current image $x$, where $N$ = imageLength $\times$ imageWidth / kernelSize$^2$. In this sense, the number of patches in different scales is depended on the image size.}

{
	\textbf{Differences with DeepEMD  \cite{Zhang_2020_CVPR}.}
	The idea of comparing patches of images using the Wasserstein distance can be also referred to other works, \eg DeepEMD.
	Concretely, DeepEMD calculates the similarity between local features of two images for few-shot learning. 
	In contrast, IWDA measures the Wasserstein distance of the distribution of patches in two images and aims to craft adversarial examples that can 
	confuse the classifier.
}

\section{Adversarial Defense with IWD} \label{sec:defense}
Relying on IWD, our proposed method (IWDA) is able to effectively attack the classifier by distribution perturbations on the manifold.
Note that the attack and defense are coupled tasks.
Essentially,
with sufficient adversarial examples, we are able to learn a robust classifier.
In this sense, we propose a new defense method with IWD, called IWDD.

	\begin{algorithm}[t]
		\caption{Defense method (IWDD).}
		\label{alg:defense}
		\begin{algorithmic}[1]\small
			\INPUT  Training data $ \{\bx_i, \by_i\}_{i=1}^n $, the number of {epochs} $ T $, the {batch} size $ m $, the hyper-parameters $ \lambda, \tau, \beta $, the number of iterations of the discriminator per generator iteration $ n_{\text{critic}} $. \\
			\OUTPUT The classifier $ h $, the discriminator $ f $, the generator $ g $,
			\FOR{ $ t= 1, \ldots, T $} 
			\STATE Sample a mini-batch $ (\bx, \by) \sim \widehat{\mD} $
			\STATE Generate adversarial examples $ \widetilde{\bx} $ using Alg. \ref{alg:attack}
			\STATE Construct a set of adversarial examples $ \mB_x(\, \epsilon(\mW)) $
			\STATE Update the discriminator and generator using Alg. \ref{alg:attack}\\
			\vspace{5pt}
			~~~~~~~~$ \nabla_{w, v} \left[ - \hat{\mmE}_{\widehat{\mD}} \left[ \mL(h(\widetilde{\bx}), \by) \right]\right], \widetilde{\bx} {\in} \mB_x(\bx, \epsilon(\mW))$
			\vspace{6pt}
			\STATE Update the classifier $ h $ by descending the gradient: \\
			\vspace{6pt}
			~~~~~~~~~ $\nabla_{\theta} \left[ \hat{\mmE}_{\widehat{\mD}} \left[ \beta \mL(h(\bx), \by)  + \mL(h(\widetilde{\bx}), \by) \right]  \right] $  \\
			\vspace{6pt}
			\ENDFOR
		\end{algorithmic}
	\end{algorithm}

% \subsection{Objective Function for Defense}
Mathematically, the \textbf{adversarial defense} problem aims to learn a robust classifier $h$ to defend against the attacks from an adversarial generator $g$ to be learned.
Following  \cite{madry2017towards, shafahi2019adversarial}, we can learn $h$ and $g$ simultaneously by solving the following minimax problem:
\begin{align}\!\!\!\!\!\min\limits_h\!\! \mathop{\mmE}\limits_{(\bx, \by) {\sim} \mD} \left[ \mathop{\max}\limits_{g} \mL(h(\widetilde{\bx})) \by) \right], \st\; \widetilde{\bx} {\in} \mB_x(\bx, \epsilon(\ell_p)).
\label{defense_AT_function}
\end{align}	

However, due to the limitations of the $\ell_p$ distance, {the Solutions like (\ref{defense_AT_function})} are limited {in improving} the robustness of a model.
By contrast, IWD is able to obtain {a} large perturbation from Theorem \ref{thm:ball_W_lp}.
{To achieve a robust model, we derive an upper bound to develop our IWDD relying on the IWD}.

To begin with, we define the expected classification error as $ \mE_{\mD}(h){=} {\mmE}_{(\bx, \by)\sim \mD} \left[ \1{ \{ h(\bx) {\neq} \by \}} \right] $ and let $\mE_{\mD}^*{=}\inf_h\mE_{\mD}(h)$. Following \cite{zhang2019theoretically}, we define the attack classification error as below.

\begin{deftn} \emph{\textbf{(Attack classification error)}}\label{def:attack_cls_error}
	Given a sample $ \bx $, if there exists an $ \epsilon(d)$-adversarial example $\widetilde{\bx} \in \mB_x(\bx, \epsilon(d))$ such that $h(\widetilde{\bx}) \neq {\by}$, the attack classification error of a classifier can be defined as:
	\begin{align*}
	\mE_{\mB_x}(h) {=} \mmE_{(\bx, \by) {\sim} \mD} \left[ \1{\left\{\exists\; \widetilde{\bx} {\in} \mB_x(\bx, \epsilon(d)), \;\st\; h(\widetilde{\bx}) {\neq} {\by} \right\}}\right].
	\end{align*}
\end{deftn}

Based on Definition \ref{def:attack_cls_error}, we derive the following upper bound to achieve the robustness of the model.

\begin{thm} \emph{\textbf{(Upper bound)}} \label{thm:upper_bound}
	Let $ \mL(h) {=} \mmE [\mL(h(\bx) \by)] $ be the $ \mL $-risk of  $ h $ and its optimum $ \mL^* {=} \inf_h \mL (h) $.
	There exists a concave function $ \xi(\cdot) $ on $ [0, \infty) $ such that $ \xi(0) {=} 0 $ and $ \xi(\delta) {\rightarrow} 0 $ as $ \delta {\rightarrow} 0^+ $
	, we have the following upper bound:
	\begin{equation*}
	\begin{aligned}
	\mE_{\mB_x} (h) {-} \mE^*_{\mD} 
	\leq  \xi (\mL(h) {-} \mL^*) 
	{+} \mmE \left[ \sup_{\widetilde{\bx} {\in} \mB_x(\bx, \epsilon(\mW))} \!\mL(h(\widetilde{\bx}) \by) \right]. \label{ineqn:upper_bound}
	\end{aligned}
	\end{equation*}
\end{thm}

From Theorem \ref{thm:upper_bound}, we minimize the upper bound such that $ \mE_{\mB_x} (h) {-} \mE^*_{\mD}  $ can be small.
In this way, the attack probability can be guaranteed to be sufficiently small.
To this end, we propose a robust minimax objective function as follows:
\begin{align} 
\!\!\!\min_{h} \!\! \mathop{\mmE}_{(\bx, \by){\sim}\mD} \Big[ \beta \mL(h(\bx) \by) {+}  \!\max_{\widetilde{\bx}{\in}\mB_x(\bx, \epsilon(\mW))}  \mL(h(\widetilde{\bx}) \by) \Big], \label{problem:defense}
\end{align}
where $ \widetilde{\bx} {=} g(\bx, \bz) $, and $\beta$ is a hyper-parameter in practice. In Problem (\ref{problem:defense}), the first term aims to minimize the natural classification loss between the prediction $ h(\bx) $ and the ground-truth $ \by $, while the second term encourages generating $\epsilon(\mW)$-adversarial examples.
By optimizing Problem (\ref{problem:defense}), the generated adversarial examples can be 
covered the data distribution to improve the robustness of the classifier.

{\textbf{Extension to multi-class problems.} }
{So far, we only focus on the binary classification problem. 
Actually, for multi-class problems, a surrogate loss is calibrated if minimizers of the surrogate risk are also minimizers of the 0-1 risk \cite{pires2016multiclass}.
The 
multi-class calibrated loss also include the cross-entropy loss \cite{zhang2019theoretically}. Therefore, we generalize Problem (\ref{problem:defense}) to the multi-class classification problems as:
\begin{align} 
\!\!\!\min_{h} \!\! \mathop{\mmE}_{(\bx, \by){\sim}\mD} \Big[ \beta \mL(h(\bx), \by) {+}  \!\max_{\widetilde{\bx}{\in}\mB_x(\bx, \epsilon(\mW))}  \mL(h(\widetilde{\bx}), \by) \Big], \label{problem:multi_defense}
\end{align}
where $\mL(\cdot)$ is the cross-entropy loss. The detailed algorithm is shown in Algorithm \ref{alg:defense}.
}

{\textbf{Differences with TRADES \cite{zhang2019theoretically}.} 
The idea of extending the binary classification to the multi-class classification was also presented in TRADES. However, the perturbation ball they defined in Problem (\ref{problem:multi_defense}) is in the $\ell_p$ distance and thus they generate adversarial examples by the PGD attack \cite{madry2017towards}. On the contrary, our IWDD algorithm uses the IWD metric to stimulate diverse adversarial examples to cover the low-density regions of the data, which would lead to better defense performance.}

	\begin{table}[t]
		\normalsize
		\centering
		\caption{Comparisons of attack success rate for different attack methods on ImageNet (3-8 lines: untargeted; 9-10 lines: targeted). {``-'' indicates the method is unsuitable for this type of attack. We report the worst-case accuracy for all the attacks.}}
		\resizebox{0.60\textwidth}{!}{
			\begin{tabular}{c|ccccccc}
				\hline
				\multirow{2}{*}{Model}        
				&  \multicolumn{6}{c}{\begin{tabular}[c]{@{}c@{}}Attack success rate (ASR) (\%) \end{tabular}} 
				\\ \cline{2-7}
				& FGSM
				& stAdv
				& EOT
				& ADef
				& FWAdv
				& IWDA \\
				\hline
				\multirow{1}{*}{Inception-v3} 
				& 69.78
				& 99.45
				& 99.73
				& \bf 100.00
				& 98.34
				& \bf 100.00 \\
				\multirow{1}{*}{Inception-v3 (free-PGD)}
				& 15.51
				& 69.31
				& 88.12
				& 97.36
				& 76.24
				& \bf100.00 \\
				\multirow{1}{*}{Inception-v3 (fast-AT)}
				& 19.17
				& 73.31
				& 88.72
				& 97.74
				& 84.21
				& \bf99.34\\
%				\hline
				\multirow{1}{*}{ResNet-101} 
				& 89.43
				& 99.73
				& 99.46
				& \bf100.00
				& \bf100.00
				& \bf100.00 \\
				\multirow{1}{*}{ResNet-101 (free-PGD)}
				& 18.83
				& 61.42
				& 91.12
				& 97.22
				& 74.58
				& \bf100.00 \\
				\multirow{1}{*}{ResNet-101 (fast-AT)}
				& 17.39
				& 60.54
				& 94.31
				& 97.32
				& 74.25
				& \bf99.62\\
%				\hline
				
				\hline \multirow{1}{*}{Inception-v3 (targeted)}
				& 0.27
				& 18.96
				& 48.63
				& 55.77
				& -
				& \bf90.66 \\
				\multirow{1}{*}{ResNet-101 (targeted)}
				& 1.90
				& 59.35
				& 79.13
				& 87.26
				& -
				& \bf94.58\\
				
				\hline
			\end{tabular}
		}
		\label{tab:untargeted_attack}
        %\vspace{-1em}
	\end{table}
	
	\begin{table}[t]
		\normalsize
		\centering
		
		\caption{Comparisons of attack success rate for different attack methods on tiny ImageNet.}
		\resizebox{0.48\textwidth}{!}{
			\begin{tabular}{c|cccc}
				\hline
				\multirow{2}{*}{Model}        
				&  \multicolumn{4}{c}{\begin{tabular}[c]{@{}c@{}}Attack success rate (ASR) (\%) \end{tabular}} 
				\\ \cline{2-5}
				& FGSM
				& ADef
				& FWAdv
				& IWDA \\
				\hline
			
				\multirow{1}{*}{ResNet-101 (PGD)}
				& 9.53
				& 79.77
				& 21.39
				& \bf91.55 \\
				\multirow{1}{*}{ResNet-101 (TRADES)}
				& 7.95
				& 75.43
				& 26.57
				& \bf92.69 \\
%				\hline
				
				\hline
			\end{tabular}
		}
		\label{tab:PGD_trades_ASR}
	\end{table}
	
\section{Experiments} \label{sec:experiments}

\textbf{Dataset.}
Following \cite{alaifari2018adef}, We conduct experiments on ImageNet \cite{olga2015imagenet}.  
In addition, we choose 10 classes from ImageNet (called tiny ImageNet), \eg barn, megalith, alp, cliff, coral reef, lakeside, seashore, valley, volcano, and coral fungus. The results on CIFAR-10 \cite{krizhevsky2009learning} are provided in Supplementary.

\textbf{Implementation details.}
We implement our method based on PyTorch and use the architectures of the models following \cite{rottshaham2019singan}. 
Specifically, we use two {pre-trained} classifiers (\ie Inception-v3 \cite{szegedy2016rethinking} and ResNet-101 \cite{he2016deep}) in the attack methods. During the training, we use an Adam optimizer to update the generator and the discriminator models with $\beta_1{=}0.5$ and $\beta_2{=}0.999$ and set the learning rate as 0.0005.
For the defense,
we use an SGD optimizer to train a new classifier with 200 epochs, a learning rate of $ 0.1 $, and {a} batch size of 128.

\textbf{Baselines and evaluation metrics.}
 We compare our attack method { with non-$\ell_p$ and $\ell_p$ attacks. Specifically, the non-$\ell_p$ attacks include
 ADef \cite{alaifari2018adef}, EOT \cite{athalye2018synthesizing}, stAdv \cite{xiao2018spatially} and FWAdv \cite{icmlWuWY20}, while the $\ell_p$ attack includes 
 FGSM \cite{goodfellow2014explaining}.} We use attack success rate (ASR)
 to measure the probability of successfully attacking the classifier.
For defense, we use free-PGD \cite{shafahi2019adversarial}, PGD \cite{madry2017towards}, fast-AT \cite{Wong2020FastIB}, TRADES \cite{zhang2019theoretically} and MART \cite{wang2019improving} as baselines. We calculate the classification accuracy on clean samples and adversarial examples.

	\begin{figure*}[t]
		\centering
		{
			\includegraphics[width=0.83\linewidth]{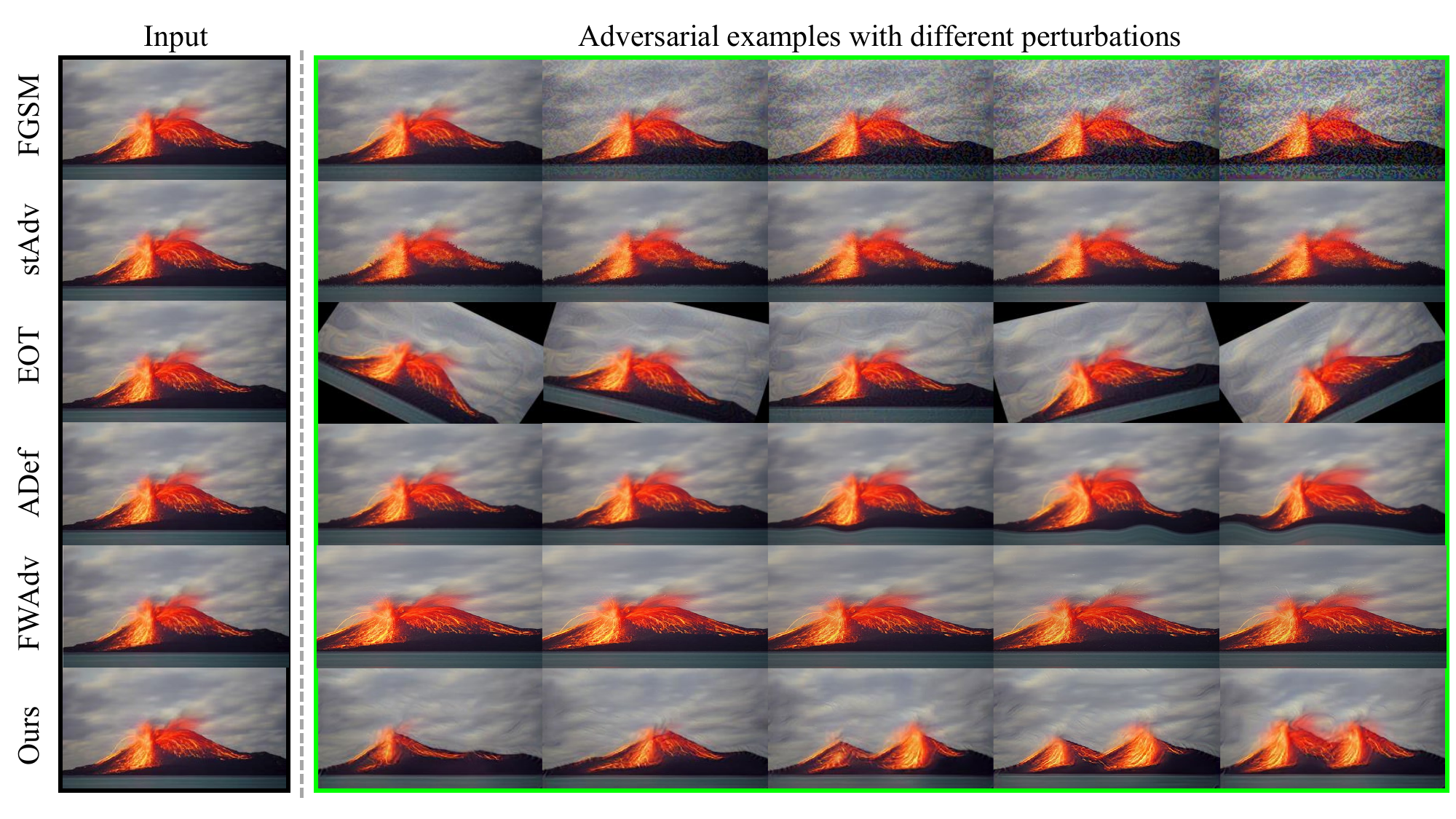}
			\vspace{-2pt}
			\caption{Comparisons of generated untargeted adversarial examples with different intensities of perturbation. (FGSM: step size; stAdv: flow; EOT: rotations; ADef: smoothness; FWAdv: mass shift; Ours: internal Wasserstein distance) }
			\label{fig:ae_imagenet}
		}
	\end{figure*}

\subsection{Experiments on Adversarial Attack} \label{sec:attack}
We perform the adversarial attack on Inception-v3 and ResNet-101 models. {We use free-PGD \cite{shafahi2019adversarial} and fast-AT \cite{Wong2020FastIB} to train robust models on ImageNet, while PGD \cite{madry2017towards} and TRADES \cite{zhang2019theoretically} on tiny ImageNet.}
Here, we use free-PGD  and fast-AT instead of PGD \cite{madry2017towards} since PGD is inefficient when training on ImageNet \cite{shafahi2019adversarial}.
Note that non-$\ell_p$ attack methods often lead to deformation, while $\ell_p$ attack methods move a few pixels in the image. Thus, it is hard to use the same distance metric to measure the similarity between samples when comparing different attack methods ($\ell_p$ and non-$\ell_p$). 
Following the settings in~\cite{Engstrom2019ExploringTL,xiao2018spatially}, we maintain the visual similarity of samples when enforcing attacks.

	\begin{table}[t]
		\normalsize
		\centering
		\caption{Accuracy (\%) of robust models on 
		tiny ImageNet. (``natural'' training uses only clean data while others use both training data and generated adversarial examples.) }
		\label{tab:defense}
		\resizebox{0.55\textwidth}{!}{
			\begin{tabular}{c|c|ccc}
				\hline
				Model      & Training method & Clean data & PGD-10 & FWAdv\\ \hline
				\multirow{4}{*}{Inception-v3} 
				& natural
				& \textcolor{blue}{83.00}
				& 0.00
				& 0.00
				\\ 
				& free-PGD      
				& 62.80 
				& 28.40
				& 54.60
				\\ 
				& PGD 
				& 70.60
				& 45.00
				& 53.20
				\\ 
				& IWDD     
				& \bf70.80
				& \bf46.00
				& \bf63.40
				\\ \hline
				\multirow{7}{*}{ResNet-101} 
				& natural
				& \textcolor{blue}{79.80}
				& 0.00
				& 0.00
				\\ 
				& free-PGD      
				& 41.40
				& 24.20
				& 36.20
				\\ 
				& PGD 
				& 69.20
				& 44.00
				& \bf54.40
				\\ 
				& fast-AT      
				& 50.80
				& 39.40
				& 44.60
				\\
				& TRADES
				& 70.00
				& 45.00
				& 51.60
				\\
				& MART
				& 62.20
				& 44.20
				& 51.20
				\\ 
				& IWDD    
				& \bf71.80   
				& \bf45.60
				& 52.00
				\\ \hline
			\end{tabular}
		}
	\end{table}

\begin{table}[t]
	\centering
	\vspace{-0.7em}
	\caption{Performance of our methods using different $\tau$ and $\beta$ on the validation set of tiny ImageNet.}
	\label{ablation_study}
	\resizebox{0.55\textwidth}{!}{
	\begin{tabular}{c|cccc}
		\hline
		$\tau$ & 0.001 & 0.01 & 0.1 & 1 \\ 
		\hline
		\multirow{1}{*}{Attack success rate (\%)} 
		& 92.41
		& 94.31
		& 100.00
		& 100.00
		
		\\ 
		\hline
		\hline
		$\beta$  & 0.001 & 0.01 & 0.1 & 1 \\ 
		\hline
		\multirow{1}{*}{Accuracy on clean data (\%)}
		& 70.20
		& 70.00
		& 71.00
		& 71.40
		\\
		\multirow{1}{*}{Accuracy on PGD-10 (\%)} 
		& 44.60
		& 45.20
		& 46.00
		& 40.00			
		\\
		\hline
	\end{tabular}
	}
% 	\vspace{-0.8em}
\end{table}

%\vspace{-5pt}
\textbf{Quantitative comparisons.}
In Table \ref{tab:untargeted_attack} and Table \ref{tab:PGD_trades_ASR}, IWDA achieves the highest ASR than other baselines.
It suggests that IWDA is able to generate more effective adversarial examples 
with the IWD.
In contrast, {the perturbation based on the $\ell_p$ by FGSM is insufficient to confuse the classifier.
Moreover, EOT, stAdv, ADef and FWAdv based on the non-$\ell_p$ apply pixel based transformations to improve ASR.} However, when attacking a robust model like free-PGD and fast-AT, these methods are difficult to achieve high ASR results since their transformations (\eg perturbation or distortion) fail to result in a large perturbation in Euclidean space. We provide more attack results in Supplementary.

\textbf{Qualitative comparisons.}
In Figure \ref{fig:ae_imagenet}, our IWDA is able to generate diverse adversarial examples by exploiting the internal distribution of the image.
These samples are far away from the input image {in Euclidean space} and hence cause a large perturbation.
Moreover, they are realistic and semantically invariant for human understanding.
In this sense, IWDA has good generalization to generate diverse adversarial examples and helps the classifier to understand a real-world image from different views.
In contrast, the baselines lack diversity although using a large perturbation to the images.
{For example, FGSM often destroys the image resolution; EOT keeps the adversarial the same as the original image but changes different rotations; stAdv damages the smoothness of edges of objects; FWAdv reflects shapes and contours in original images but also disrupts the image resolution; ADef maintains the smoothness of the image but deforms the original image.}
{Besides, we show the qualitative results of targeted attack in Supplementary.}

\begin{figure}[t]
		\centering
		{	\includegraphics[width=0.95\linewidth]{ 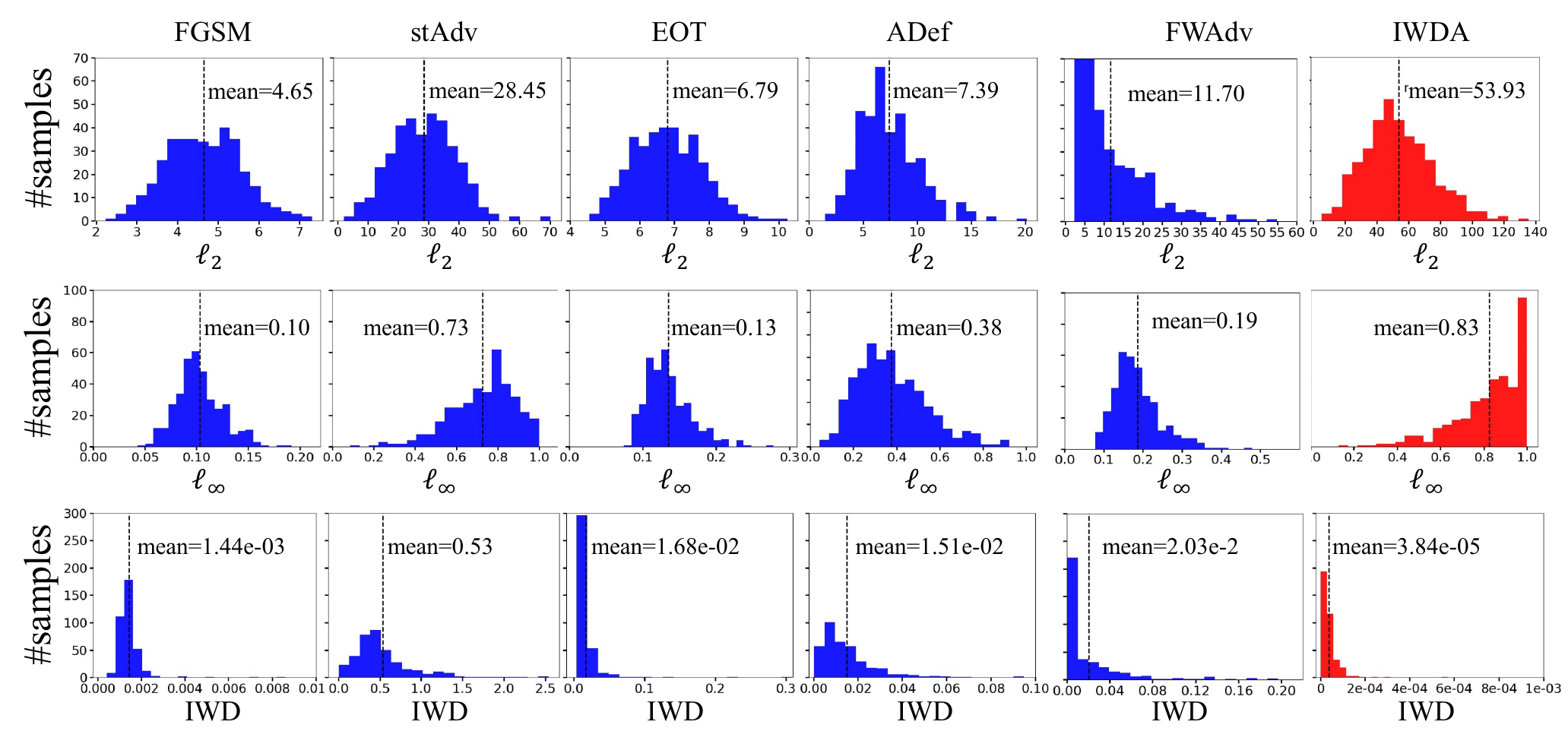}
		\vspace{-4pt}
		\caption{Comparisons of the perturbation for different methods. We use {the} $\ell_2$ distance, {the} $\ell_{\infty}$ distance and IWD to measure the distance between the original samples and perturbed samples.}	\label{fig:l2_distance}
		}
		 \vspace{-0.8em}
	\end{figure}

\textbf{Comparisons of perturbation.}
We further compare the perturbation distance of different methods. Under the untargeted attack setting, we use {pre-trained} ResNet-101 to generate adversarial examples on tiny ImageNet. 
Specifically, we measure three metrics (\ie $\ell_2$, $\ell_\infty$ and IWD)  between a sample and its adversarial example. We count the number of perturbed samples over three metrics. {In the first two lines of Figure~\ref{fig:l2_distance}, the adversarial examples using the IWDA 
achieve the highest mean over the $\ell_2$ and $\ell_\infty$ in Euclidean space compared to baseline methods, thus it has larger perturbations to attack the model.
More importantly, in the last line, IWDA achieves the smallest IWD values, even close to zero, suggesting that IWDA is powerful in attacking while keeping the high visual similarity of samples.}

\subsection{Experiments on Adversarial Defense} %\label{sec:defense}
We use Inception-v3 and ResNet-101 to demonstrate the robustness of IWDD.
We train all defense methods on tiny ImageNet and use PGD-10 \cite{madry2017towards} and FWAdv \cite{icmlWuWY20} to attack the mdoels. We present the clean accuracy and the adversarial accuracy.
In Table \ref{tab:defense},	
IWDD achieves a superior or comparable robustness compared to defense baselines. Only against FWAdv on ResNet-101, PGD achieves better performance than IWDD. We believe this is an inherited trade-off between clean accuracy and adversarial accuracy \cite{zhang2019theoretically}. Other defense baselines like TRADES and MART are limited to improving the generalization on both clean accuracy and adversarial accuracy. Moreover, we put more defense results under AutoAttack \cite{croce2020reliable} in Supplementary.

\subsection{Ablation Study} \label{sec:abaltion}
We further evaluate the influences of the hyper-parameters $ \tau $ in Eqn.~(\ref{problem:total}) and $ \beta $ in Eqn.~(\ref{problem:multi_defense}), respectively.
By setting different values, we attack the {pre-trained} ResNet-101  for IWDA and train a robust ResNet-101
for IWDD. The results of attack success rate and validation accuracy 
are put in Table \ref{ablation_study}.
Empirically, when 
setting $\tau{=}0.1$ and $\beta{=}0.1$, IWDA and IWDD achieve superior performance on the trade-off between clean accuracy and robust accuracy.

\section{Conclusion} \label{sec:conclusion}
In this paper, we have proposed an internal Wasserstein distance (IWD) 
to measure the similarity between two samples.
Relying on IWD, we propose a novel attack method to generate diverse adversarial examples.
Meanwhile, we derive an upper bound and propose a new defense method to improve the robustness of the classifier.
Moreover, we analyze the generalization performance of our defense method and prove that robustness requires more diverse adversarial examples in the adversarial training. 
Extensive experiments on ImageNet confirm the effectiveness of our both attack and defense methods. Furthermore, we expect that a good distance metric will encourage a plethora of researchers to pursue similar tasks in the future.

\section*{Acknowledgments}
This work was partially supported
by Key-Area Research and Development Program of
Guangdong Province (2018B010107001, 2019B010155002,
2019B010155001), National Natural Science Foundation
of China (NSFC) 61836003 (key project), National Natural Science Foundation of China (NSFC) 62072190,
2017ZT07X183, Tencent AI Lab Rhino-Bird Focused Research Program (No.JR201902), Fundamental Research
Funds for the Central Universities D2191240.
%Bibliography
\bibliographystyle{unsrt}  
\bibliography{IWD}  

\newpage
\begin{table}[h]
	\begin{tabular}{p{1\columnwidth}}
		\centering
		\Large{\textbf{Supplementary Materials: Learning Defense Transformers  for Counterattacking Adversarial Examples}}
% 		\vspace{3pt}
	\end{tabular}
\end{table}

\setcounter{equation}{0}
\setcounter{table}{0}
\setcounter{figure}{0}

\setcounter{section}{0}
\renewcommand\thesection{\Alph{section}}

We organize our Supplementary  as follows.
In Section \ref{supp_sec:proof_attack_prob}, we prove the Theorem \ref{thm:ball_W_lp} that IWD can result a larger perturbation than the $\ell_p$ distance.
In Section \ref{supp_sec:proof_upper_bound}, we prove the Theorem \ref{thm:upper_bound} about the upper bound.
In Section \ref{supp_sec:implementation_details}, we give the implementation details for both attack and defense methods.
In Section \ref{supp_sec:more_results}, we present more experimental results for our proposed methods.

\section{Proofs of Theorem \ref{thm:ball_W_lp}} \label{supp_sec:proof_attack_prob}

\begin{proof}
    %We first prove that the $\ell_{\infty}$ attack can lead to the largest perturbation ball in the $\ell_p(p >0)$ distance, while it is in fact a specific case of the IWD attack. Then, we discuss some cases that the IWD attack can obtain a larger perturbation ball than the $\ell_{\infty}$ attack.
    Based on the definition of diameter of a set $\mA$, \ie
    \begin{align*}
        \diam (\mA) := \sup_{\bx', \bx'' \in \mA} d(\bx', \bx''),
    \end{align*}
    we focus on the maximum distance of the set $\mA$.
    For convenience of analysis, we use the $\ell_1$ distance to measure the similarity between any $\bx'$ and $\bx''$ in the set $\mA$.
    
    For any $\epsilon(\ell_p)>0$ to $\ell_p$ attack methods, the diameter of the $\mB_x(\bx, \epsilon(\ell_p))$ satisfies:
    \begin{align}
        \label{supp_fomular:dim_ell_p}
        \diam (\mB_x(\bx, \epsilon(\ell_p))) &= \sup_{\bx', \bx'' \in \mB_x(\bx, \epsilon(\ell_p))} d(\bx', \bx'') \leq  \sup_{\widetilde{\bx} \in \mB_x(\bx, \epsilon(\ell_p))} 2\sum_{i=1}^m \|\bx_i - \widetilde{\bx}_{i} \|_1 \nonumber\\
        &\leq  2{m} \epsilon(\ell_p) \leq 2m\epsilon(\ell_{\infty}),
    \end{align}
    where $\bx_i, \widetilde{\bx}_i \in \mB_x(\bx, \epsilon(\ell_p))$ are patches of $\bx$ and $\widetilde{\bx}$,
    $m$ is the dimension of the image $\bx$. The last inequality follows a fact that  the $\ell_{\infty}$ attack, \ie{$\widetilde{\bx} = \bx \pm \epsilon(\ell_p)$} can lead to the largest perturbation ball in (\ref{supp_fomular:dim_ell_p}) in the $\ell_{\infty}$ attacks. Therefore, we next only compare the  $\ell_{\infty}$ attack and IWD attack.%and $\diam (\mB_x(\bx, \epsilon(\ell_p))) \small{=} 2\sqrt {m} \epsilon(\ell_p)$ happens only when $\widetilde{\bx} = \bx \pm \epsilon(\ell_p)$ in the $\ell_{\infty}$ attack.

    Based on the definition of the internal Wasserstein distance, 
	\begin{align*}
		%\label{def:IWD}
		\mW(\bx, \widetilde{\bx}) := \inf_{\gamma \in \Gamma(\mu, \nu)} \mmE_{(\bu, \bv) \sim \gamma} \left[ \| \bu - \bv \|_1 \right],
	\end{align*}
    where $\mu$ and $\nu$ are internal distributions of $\bx$ and $\widetilde{\bx}$, respectively.
    
    Note that when the patch size of the $\bx$ is scaled down to one, for the case $\widetilde{\bx} = \bx \pm \epsilon(\ell_{\infty})$, we have
    \begin{align}
        \label{supp_def:IWD_upper}
        \mW(\bx,\widetilde{\bx}) 
        % &= \inf_{\gamma \in  \Gamma(\mu, \nu)}\mmE_{(\bu, \bv) \sim \gamma} \left[ \| \bu - \bv \|_1 \right] 
        = \inf_{\gamma \in  \Gamma(\mu, \nu)}\mmE_{(\bx_i, \widetilde{\bx}_j) \sim \gamma} \left[\| \bx_i - \widetilde{\bx}_j \|_1\right] 
        \leq \mmE_{(\bx_i,\widetilde{\bx}_i) \sim  \gamma}\left[\|\bx_i - \widetilde{\bx}_i\|_1 \right]\small{=} \epsilon(\ell_{\infty}).
    \end{align}
    The result (\ref{supp_def:IWD_upper}) implies $\ell_{\infty}$ attack can be considered as one of the specific cases of the IWD attack.
    
    Next, we discuss the cases that the patch size is grater one. Note that the size of patch is smaller than the orginal image.  We can reconstruct the patches of image to craft the adversarial example in the region of $\mB_x(\bx, \epsilon(\mW))$. Without loss of generality, we partition an image $\bx$ into two equal parts, \ie $\bx = [\bx^{(1)}, \bx^{(2)}]$, and assume that $\| \bx^{(1)} - \bx^{(2)} \|_1 \geq \frac{m}{2} \epsilon(\ell_{\infty})$. Then, we suppose the adversarial example is disrupted the order of these two block, \ie $\widetilde{\bx} = [\bx^{(2)}, \bx^{(1)}]$.    Relying on the definition of the internal Wassersein distance, we have
    \begin{align*}
        \mW(\bx, \widetilde{\bx}) = 0 < \epsilon(\ell_{\infty}).
    \end{align*}
    However, the distance between $\bx$ and $\widetilde{\bx}$ satisfies that 
    \begin{align}
    \label{supp_fomular:dim_IWD}
    \diam (\mB_x(\bx, \epsilon(\mW))) &= \sup_{\bx', \bx'' \in \mB_x(\bx, \epsilon(\mW))} d(\bx', \bx'') \nonumber\\
    &\ge 2\| \bx - \widetilde{\bx} \|_1\nonumber\\
    &= 2\| \bx^{(1)} - \bx^{(2)} \|_1 + 2\| \bx^{(2)} - \bx^{(1)} \|_1 \nonumber\\
    &\ge 2m\epsilon(\ell_{\infty}).    
    \end{align}
    Combine formulas \ref{supp_fomular:dim_ell_p} and \ref{supp_fomular:dim_IWD} to achieve
    \begin{align}
        \diam (\mB_x(\bx, \epsilon(\ell_p))) \leq \diam (\mB_x(\bx, \epsilon(\mW))).
    \end{align}
\end{proof}

% \newpage
\section{Proofs of Theorem \ref{thm:upper_bound}}\label{supp_sec:proof_upper_bound}

To develop our theoretical analysis, we first give the definition of classification boundary term and give the following lemma.

\begin{lemma}\label{supp_lemma:EB_decom}
	Let the classification boundary term be defined as $ \mE_{\text{bd}}(h) := \mmE_{(\bx, \by)\sim \mD} [\1{\left\{ \bx \in \mB_h(h, \epsilon(d)), h(\bx) = \by \right\}}] $ {and the expected classification error as $ \mE_{\mD}(h){=} {\mmE}_{(\bx, \by)\sim \mD} \left[ \1{ \{ h(\bx) {\neq} \by \}} \right] $}.
	The robust classification error can be decomposed into 
	\begin{align}
		\mE_{\mB_x}(h) = \mE_{\mD}(h) + \mE_{\text{bd}}(h, \epsilon(d)).
	\end{align}
\end{lemma}
\begin{proof}
	When being far away from the classification boundary, if the sample is classified correctly (\ie $h(\bx)=\by$), its perturbation is also correct (\ie $h(\widetilde{\bx})=\by$).
	On the other hand, if the sample is misclassified (\ie $h(\bx)\neq\by$), then its adversarial example is also misclassified (\ie $h(\widetilde{\bx})\neq\by$), then $\mE_{\mB_x}(h) = \mE_{\mD}(h)$.
% 	\begin{align}
% 		\mE_{\mB_x}(h) = \mE_{\mD}(h).
% 	\end{align}
	When within the classification boundary, the sample is classified correctly, while its perturbation is misclassified, then $\mE_{\mB_x}(h) > \mE_{\mD}(h)$.
% 	\begin{align}
% 		\mE_{\mB_x}(h) > \mE_{\mD}(h).
% 	\end{align}
	Moreover, the difference between these two terms is $ \mE_{\text{bd}}(h, \epsilon(d)) = \mE_{\mB_x}(h) - \mE_{\mD}(h) $.
\end{proof}

Based on the definition of classification boundary term, we derive the following lemma to develop Theorem \ref{thm:upper_bound}. 
\begin{lemma}\label{supp_thm:existence_ae}
% \textbf{(Existence of $\epsilon(d)$-adversarial example)}
	Let $ \mL(h) = \mmE [\mL(h(\bx) \by)] $ be the $ \mL $-risk of a function $ h $ and its optimum $ \mE_{\mD}^* = \min_h \mE_{\mD} (h) $.
	There exists a concave function $ \xi(\cdot) $ on $ [0, \infty) $ such that $ \xi(0) = 0 $ and $ \xi(\delta) \rightarrow 0 $ as $ \delta \rightarrow 0^+ $, we have
	\begin{equation}
		\begin{aligned}
		\mE_{\mB_x} (h) - \mE^*_{\mD} 
		\leq\;  \xi (\mL(h) - \mL^*) + \mE_{\text{bd}}(h, \epsilon(d)).
		\end{aligned}
	\end{equation}
\end{lemma} 
\begin{proof}
	{For the binary classification,}
	the prediction $ h(\bx) $ is a vector function, and each element of $ h(\bx) $ is in $ \{-1, 1\} $.
	According to \cite{gao2011consistency}, the surrogate loss can be defined as 
	\begin{align*}
		\mL(h(\bx) \by) = \frac{1}{K} \sum_{i=1}^K \mL(h_i(\bx) y_i),
	\end{align*}
	where $ \mL $ is a convex function, $K$ is the number of labels, $h_i(\bx)$ is the $i$-th element of $h(\bx)$ and $y_i$ is the $i$-th element of $\by$. 
	For example, the hinge loss $ \mL(t) = (1-t)_+ $, exponential loss $ \mL(t) = \exp(-t) $, least squares loss $ \mL(t) = (1-t)^2 $ and logistic regression loss $ \mL(t) = \log_2 (1+\exp(-t)) $.
	Based on Lemma \ref{supp_lemma:EB_decom}, we have
	\begin{align*}
		\mE_{\mB_x} (h) - \mE^*_{\mD} 
		=& \mE_{\mD}(h) - \mE^*_{\mD} + \mE_{\text{bd}}(h, \epsilon(d)) \\
		\leq& \xi (\mL(h) - \mL^*) + \mE_{\text{bd}}(h, \epsilon(d)) \\
		=& \xi (\mL(h) - \mL^*) + \Pr \left[ \bx \in \mB_h(h, \epsilon(d)), h(\bx) = \by \right].
	\end{align*}
	The second line follows by Corollary 26 of \cite{zhang2004statistical}, it guarantees there exists a concave function $ \xi $ on $ [0, \infty] $ such that $ \xi(0)=0 $ and $ \xi(\delta) \rightarrow 0 $ as $ \delta \rightarrow 0^+ $ and 
	\begin{align*}
	    \mE_{\mD}(h) - \mE^*_{\mD} \leq \xi (\mL(h) - \mL^*).
	\end{align*}
\end{proof}

\newpage
\textbf{Theorem \ref{thm:upper_bound} (Upper bound)}
\emph{
	Let $ \mL(h) = \mmE [\mL(h(\bx) \by)] $ be the $ \mL $-risk of a function $ h $ and its optimum $ \mL^* = \min_h \mL (h) $.
	There exists a concave function $ \xi(\cdot) $ on $ [0, \infty) $ such that $ \xi(0) = 0 $ and $ \xi(\delta) \rightarrow 0 $ as $ \delta \rightarrow 0^+ $, we have
	\begin{align}
    	\mE_{\mB_x} (h) - \mE^*_{\mD} 
    	\leq\;  \xi (\mL(h) - \mL^*) + \mmE \left[ \sup_{\widetilde{\bx} \in \mB_x(\bx, \epsilon(\mW))} \mL(h(\widetilde{\bx}) \by) \right]. 
	\end{align}
}
\begin{proof}
Relying on the conditions of Theorem \ref{supp_thm:existence_ae}, we have the following upper bound,
	\begin{align*}
		~~~~~~~~~&~~~~\mE_{\mB_x} (h) - \mE^*_{\mD} \\
		\leq& \xi (\mL(h) - \mL^*) + \mE_{\text{bd}}(h, \epsilon(d))\\
		=& \xi (\mL(h) - \mL^*) + \Pr \left[ \bx \in \mB_h(h, \epsilon(\mW)), h(\bx) = \by \right]\\
% 		\leq& \xi (\mL(h) - \mL^*) + \Pr \left[ \bx \in \mB_x(\bx, \epsilon(\mW)), h(\bx) = \by \right]\\
        =& \xi (\mL(h) - \mL^*) + \mmE \left[  \1 {\left\{ 
        \left\{ \bx {\in} \mB_h(h, \epsilon(\mW)),h(\bx) = \by \right\}
        \wedge 
        \left\{ \exists\; \widetilde{\bx} {\in} \mB_x(\bx, \epsilon(\mW)),
        h(\widetilde{\bx}) \neq h(\bx) \right\}
        \right\}} \right] \\
		\leq& \xi (\mL(h) - \mL^*) + \mmE \left[ \sup\limits_{\widetilde{\bx} \in \mB_x(\bx, \epsilon(\mW))} \1{\left\{ h(\widetilde{\bx}) \neq \by \right\}} \right] \\
		=& \xi (\mL(h) - \mL^*) + \mmE \left[ \sup\limits_{\widetilde{\bx} \in \mB_x(\bx, \epsilon(\mW))} \1{\left\{ h(\widetilde{\bx}) \cdot \by < \0 \right\}} \right] \\
		\leq& \xi (\mL(h) - \mL^*) + \mmE \left[ \sup\limits_{\widetilde{\bx} \in \mB_x(\bx, \epsilon(\mW))} \frac{1}{K} \sum\limits_{i=1}^K \mL \left( h_i(\widetilde{\bx}) h_i(\bx) \right)  \right] \\
		=& \xi (\mL(h) - \mL^*) + \mmE \left[ \sup\limits_{\widetilde{\bx} \in \mB_x(\bx, \epsilon(\mW))} \mL(h(\widetilde{\bx}) \by) \right],
	\end{align*}
	where the first two lines follows by Theorem \ref{supp_thm:existence_ae}.
	The third line is based on the definition of $\mB_h(h, \epsilon(\mW))$ and $\Pr \left[ \bx \in \mB_h(h, \epsilon(\mW)), h(\bx) = \by \right]$.
	The fourth line follows by a fact that $\mB_x(\bx, \epsilon(\mW))$ contains more perturbed samples than $\mB_h(h, \epsilon(\mW))$.
	\begin{align*}
	    &\mmE \left[  \1 {\left\{ \left\{ h(\widetilde{\bx}) \neq h(\bx) \right\} \wedge \left\{ h(\bx) = \by \right\}, \bx {\in} \mB_h(h, \epsilon(\mW)), \widetilde{\bx} {\in} \mB_x(\bx, \epsilon(\mW))\right\}} \right]\\ 
	    \leq& \mmE \left[ \sup\limits_{\widetilde{\bx} \in \mB_x(\bx, \epsilon(\mW))} \1{\left\{ h(\widetilde{\bx}) \neq \by \right\}} \right].
	\end{align*}
	The fifth line satisfies because $h(\widetilde{\bx})$ and $\by$ are valued in $\{-1, 1\}$.
	The last two lines hold by the definition of the surrogate loss \cite{gao2011consistency}, \ie
    \begin{align*}
    	\mL(h(\widetilde{\bx}), \by) = \frac{1}{K} \sum_{i=1}^K \mL(h_i(\widetilde{\bx}) y_i).
    \end{align*}
This loss includes the hinge loss $ \mL(\alpha) = (1-\alpha)_+ $, exponential loss $ \mL(\alpha) = \exp(-\alpha) $, least squares loss $ \mL(\alpha) = (1-\alpha)^2 $ and logistic regression loss $ \mL(\alpha) = \log_2 (1+\exp(-\alpha)) $, etc.
These losses have a property that 
\begin{align*}
    \mL(\alpha) \ge 1 \quad \text{when} \quad \alpha \leq 0. 
\end{align*}
Therefore, we have
\begin{align*}
    \1{\left\{ h(\widetilde{\bx}) \cdot \by < \0 \right\}} \leq \frac{1}{K} \sum\limits_{i=1}^K \mL \left( h_i(\widetilde{\bx}) h_i(\bx) \right).
\end{align*}
\end{proof}

\newpage
\section{More Implementation Details} \label{supp_sec:implementation_details}

\subsection{Details of Attack}
In our implementation, We use a natural training method, a free-PGD and fast-AT training methods to train the classifiers (Inception-v3 and ResNet-101) on ImageNet.
The test accuracies on tiny ImageNet are in Table \ref{supp_tab:acc_pretrain_classifier}.
In the following, we introduce the detailed settings for the attack.

\textbf{Untargeted attack.}
we train the GAN model with different scaled images to generate adversarial examples. Specifically, we set the scale factor $s{=}0.75$, the minimum height $H_{min}{=}25$ and the minimum size of the patches $P_{min}{=}3$. For each scale, we train each GAN model 2000 times. To perform IWDA,
we first use a set of noise vectors and an original image as inputs of the generator $g_0$, and use the original image as the input of the discriminator $f_0$. 
To achieve better performance, we use a PGD-10 adversarial example as the input of the discriminator. 
Then, we introduce a classifier at the last scale of updating the GAN model. 
Actually, we are able to attack the classifier at any scale.

In the optimization, we maximize the cross entropy loss with $\tau = 0.1$ to generate adversarial examples. 
At each iteration, we update the generator and discriminator with a step of 3.
We use an Adam optimizer to optimize the parameters of the GAN model. 
We decay the learning rate of optimizer with a milestones of 1600 and a gamma of 0.1 at each training process. Other experiments setup are the same as the untargeted attack.

% \vspace{-5pt}
\begin{table*}[t]
\renewcommand\thetable{A}
	 \vspace{-4pt}
	 \normalsize
	\centering
	\caption{Accuracy of the classifiers on the validation set of tiny ImageNet.}
	\resizebox{1\textwidth}{!}{
		\begin{tabular}{c|ccc|ccc}
			\hline
			Classifier
			& Inception-v3
			& Inception-v3 (free-PGD)
			& Inception-v3 (fast-AT)
			& ResNet-101
			& ResNet-101 (free-PGD) 
			& ResNet-101 (fast-AT) \\
			\hline
			Accuracy (\%) 
			& 72.80
			& 60.60
			& 59.80
			& 73.80
			& 64.80
			& 53.20\\
			\hline
		\end{tabular}
	}
	\label{supp_tab:acc_pretrain_classifier}
	\vspace{-0.8em}
\end{table*}

\textbf{Targeted attack.}
We further conduct targeted attack experiments with random target labels. 
In contrast to the untargeted attack setting, we perform the targeted attack at each scale. %$s = 0$. 
In the optimization, we minimize the cross entropy loss with $\tau = 0.01$. 
First, we train each GAN model with 2000 iterations. 
At each iteration, we update generators and discriminators with 3 steps. 
Then, we update the generators with the cross entropy loss only at the first step. 
We use the same optimizer settings as the untargeted attack.

{\textbf{Detailed settings for attack methods}.
We conduct experiments of adversarial attack 
with the following parameter settings. For FGSM \cite{goodfellow2014explaining}, we let $\epsilon = 2/255$ be the magnitude of the perturbation; for stAdv \cite{xiao2018spatially}, we set $\tau = 0.05$; for EOT \cite{athalye2018synthesizing}, we set the rotation range
at $[-45^{\circ}, 45^{\circ}]$; for ADef \cite{alaifari2018adef}, we set $\sigma = 0.5$, overshot $= 1.2$ and iterations $T = 100$; for FWAdv \cite{icmlWuWY20}, we use the PGD Dual Proj algorithm and set $\epsilon = 0.005$ to obtain the worst attack.} 

\subsection{Details of Defense}

\textbf{Training of the GAN model.} For the proposed defense method, we use our attack method by updating the generators and discriminators to generate more realistic and diverse adversarial examples.
However, the training cost is very expensive.
To address this, we pre-train the GAN model to prepare more diverse adversarial examples for adversarial training.
Meanwhile, we train a universal generator to generate adversarial examples for efficiency.
Following the settings of \cite{xiao2018generating}, we pre-process the size of input images as 224 $\times$ 224.
Then, we use an Adam optimizer to train the GAN models with 200 epochs, a learning rate of 0.001, $\beta_1=0.5$ and $\beta_2=0.999$ to optimize the parameters of GAN models. 
Moreover, we decay the learning rate of optimizer with $10^{-1}$ per 50 epochs.
{We conduct this experiment on two NVIDIA Titan Xp GPUs.}

\textbf{Training of the classifier.}
We further conduct adversarial training experiments to train the classifiers (\ie Inception-v3 and ResNet-101). 
First, we pre-process all input images as $299 \times 299$ for Inception-v3 and $224 \times 224$ for ResNet-101, respectively.
Following \cite{alaifari2018adef}, we perturb the input images of the generator using the PGD algorithm before each iteration of the training process.
To optimize Loss (\ref{problem:defense}), we set $\beta=0.1$. 
% Besides, we set the random seed as 2.
Specifically, we train the classifier with 200 epochs.
We use an SGD optimizer with a learning rate of 0.1, a momentum of 0.9 and a weight decay of 0.0005, and we decay the learning rate of optimizer with $10^{-1}$ at epoch $=100$ and epoch $=150$, respectively. {We conduct this experiment on four NVIDIA Titan Xp GPUs.}

\textbf{Testing of the defense method.}
We finally test our robust model and other adversarial training models on both clean data and adversarial examples generated by PGD and FWAdv, as shown in Table~\ref{tab:defense}. Following ~\cite{madry2017towards}, we use the step size of $2/255$ and the clip epsilon of $8/255$ to generate the adversarial examples in PGD attack. For FWAdv attack, we use the PGD Dual Proj algorithm and set $\epsilon = 0.005$ to obtain  the adversarial examples {following \cite{icmlWuWY20}.}

% \newpage
\section{More Experimental Results} \label{supp_sec:more_results}

\subsection{More Quatitative Results of IWDA}

{\textbf{Comparisons with attack methods on CIFAR-10.}}
We also compare our IWDA with other attack methods under the untargeted attack on CIFAR-10 over ResNet18 and demonstrate the attack success rate (ASR) in Table~\ref{tab:cifar}. IWDA achieves the higher ASR than other baseline methods, which is consistent with the results on ImageNet in Table~\ref{tab:untargeted_attack}.

\begin{table}[tbp]
\vspace{-11pt}
  \centering
  \renewcommand\thetable{B}
  \caption{Comparisons of attack success rate (\%) on CIFAR-10.}
 \resizebox{0.66\textwidth}{!}
 {
    \begin{tabular}{c|cccccc}
\hline
Model          & FGSM  & stAdv & EOT   & ADef  & FWAdv & IWDA  \\ \hline
Resnet-18      & 54.18 & 95.83 & 85.13 & 99.16 &  97.33     & \bf99.89 \\
Resnet-18(PGD) & 37.89 & 41.53 & 54.38 & 97.44 &  38.19     & \bf98.42 \\ \hline
\end{tabular}
    }
    \label{tab:cifar}
    % \vspace{-5pt}
\end{table}

{\textbf{Robustness of IWDD against IWDA.}
We use IWDA method to attack the IWDD model and provide the classification accuracy of the IWDD model. From Table \ref{tab:robust_IWDD_to_IWDA}, the IWDD model achieves 5.40\% and 6.00\% classification accuracy for ResNet-101 and Inception-v3, respectively. We believe training a universal generator to craft adversarial examples for training in IWDD is efficient and effective to defend against attacks such as PGD and FWAdv in Table \ref{tab:defense}, while it is still difficult to defend against the adversarial examples crafted with IWDA. This phenomenon stems from the power of IWDA elaborately crafting adversarial examples to each sample with a  tailored generative model. We postulate that the key to addressing this robustness is training a universal generator that is  able to generate tailored adversarial examples for each sample with some conditional information, \eg the label embedding. %\cite{Mirza2014ConditionalGA}. 
We hope that future research will investigate this robustness problem.}

	\begin{table}[tbp]
		% \vspace{-10pt}
		\centering
		 \renewcommand\thetable{C}
		\caption{ Accuracy (\%) of IWDD model under the IWDA attack on tiny ImageNet for ResNet-101 and Inception-v3.}
		\resizebox{0.36\textwidth}{!}{
			\begin{tabular}{cc|cc}
			\hline
			\multicolumn{2}{c|}{Model}           & PGD-10 & IWDA  \\ 
			\hline
			\multirow{2}{*}{ResNet-101}   & PGD  & 44.00  & 5.00  \\
            & IWDD & 45.60  & 5.40  \\ 
            \hline
            \multirow{2}{*}{Inception-v3} & PGD  & 45.00  & 5.60  \\
            & IWDD & 46.00  & 6.00  \\
\hline
\end{tabular}
		}
		\label{tab:robust_IWDD_to_IWDA}
	\vspace{-8pt}
	\end{table}

\subsection{More Qualitative Results of IWDA}

\begin{figure*}[h]
\renewcommand\thefigure{A}
	\centering
	{
		\includegraphics[width=0.95\linewidth]{ 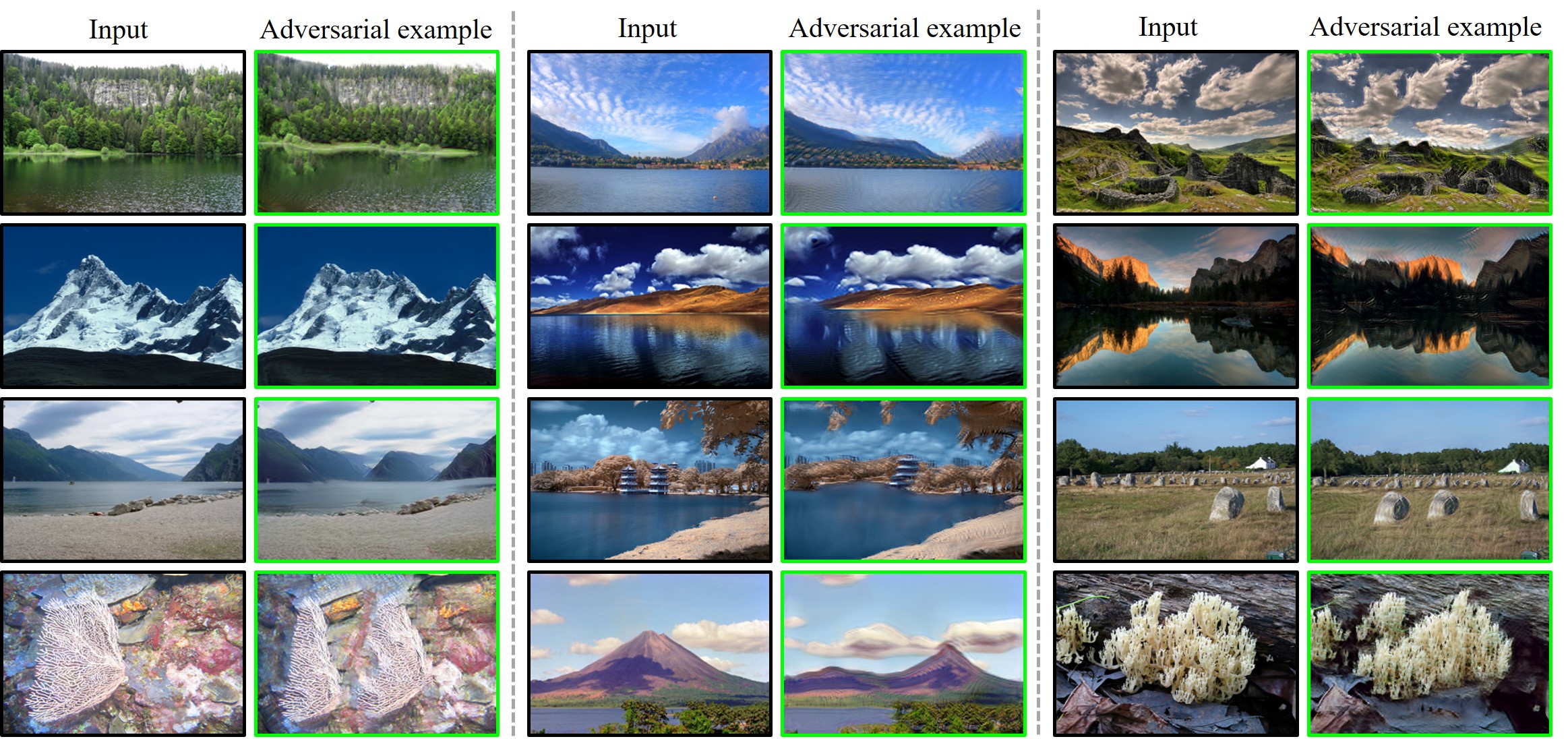}
% 		\vspace{-8pt}
		\caption{More untargeted attack results of our IWDA on Inception-v3.
		}
		\label{fig:untarget_ae_inception_supp}
	}
\end{figure*}

\begin{figure*}[h]
\renewcommand\thefigure{B}
    \centering
	{
		\includegraphics[width=0.95\linewidth]{ 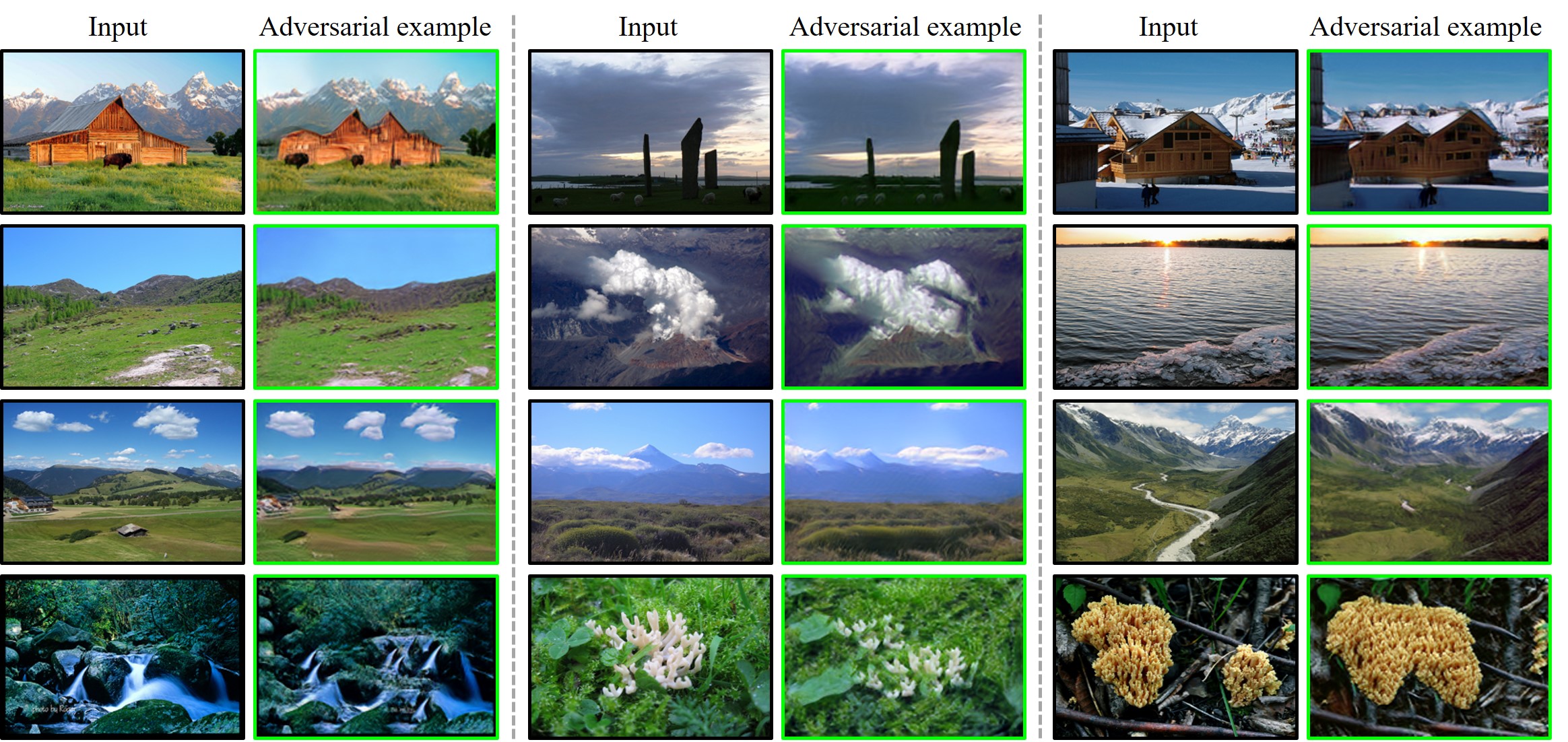}
% 		\vspace{-8pt}
		\caption{More qualitative results of our IWDA on ResNet-101 under the untargeted attack setting.
		}
		\label{fig:untarget_ae_resnet_supp}
	}
\end{figure*}

\begin{figure*}[h]
\renewcommand\thefigure{C}
	\centering
	{
		\includegraphics[width=0.95\linewidth]{ 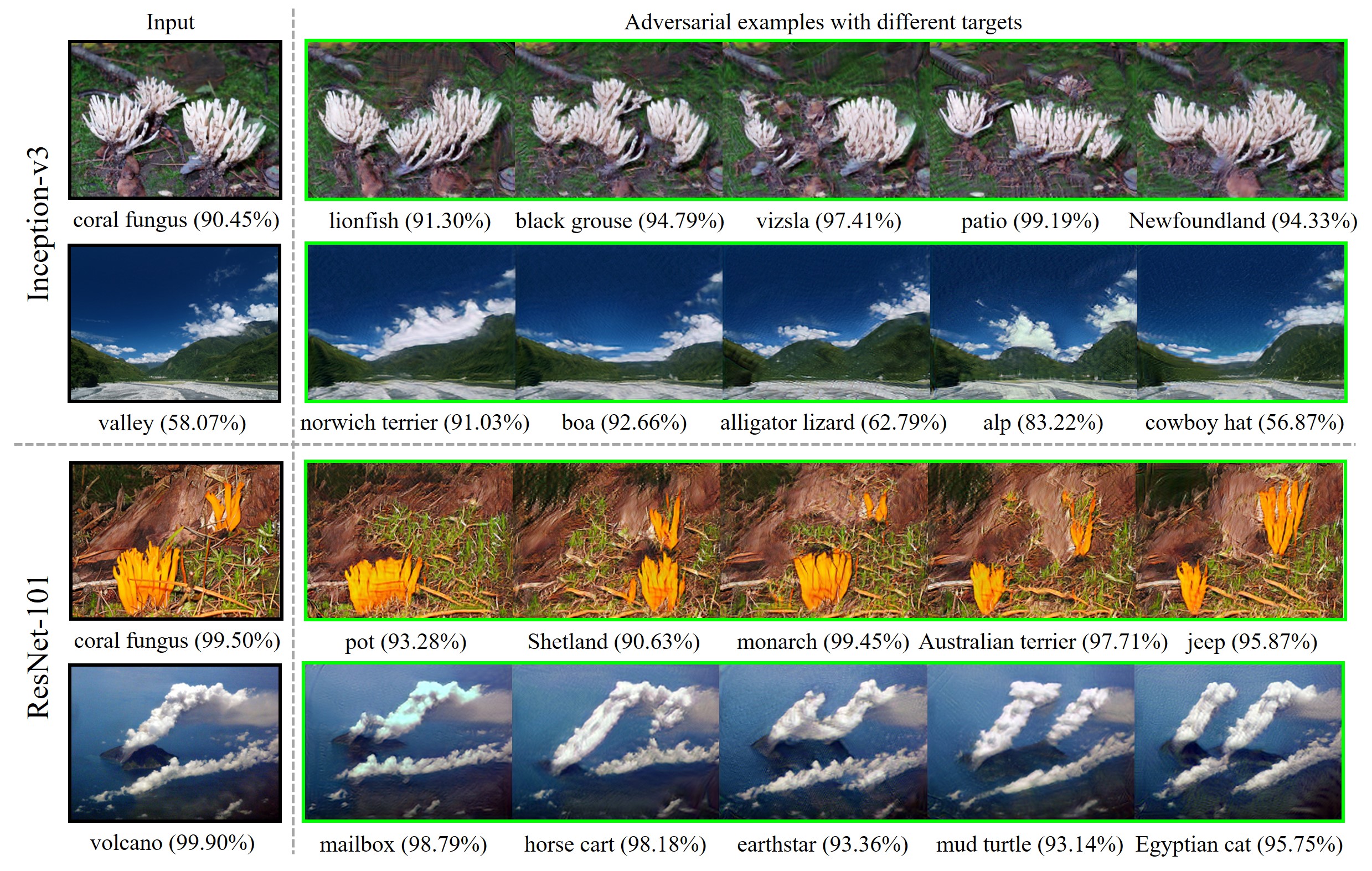}
% 		\vspace{-8pt}
		\caption{More qualitative results of our IWDA under the targeted attack setting.
		}
		\label{fig:target_ae_supp}
	}
	
\end{figure*}

%\textbf{Untargeted attack.} 
We perform our untargeted attack with the pre-trained Inception-v3  and ResNet-10 on ImageNet and present more qualitative results in Figure \ref{fig:untarget_ae_inception_supp} and Figure \ref{fig:untarget_ae_resnet_supp}, respectively. Moreover, we perform the targeted attack in Fig \ref{fig:target_ae_supp}.
These results suggest that our IWDA has good generalization performance to generate diverse and real adversarial examples and helps the classifier to understand a real-world image from different views.

\subsection{More Quatitative Results of IWDD}
To further evaluate the robustness of our IWDD, we compare with other baselines under AutoAttack \cite{croce2020reliable}. In Table~\ref{Tab:auto}, IWDD achieves superior or comparable robustness.

\begin{table}[t]
\renewcommand\thetable{D}
\vspace{-8pt}
\centering
\caption{  Robust accuracy (\%) on tiny ImageNet for ResNet-101.}
\resizebox{0.7\textwidth}{!}{
\begin{tabular}{c|cccccc}
\hline
Defense & Clean data     & APGD$_{CE}$       & APGD$_T$   & FAB$_T$         & Square         & AA             \\ \hline
PGD             & 69.20          & 42.60          & 40.80     & {42.00} & \textbf{50.40}          & 40.80          \\
TRADES          & 70.00          & 44.00          & 41.60     & \textbf{45.00}          & 47.60          & 41.60          \\
MART            & 62.20          & 43.60          &     40.20      & 41.40          &       45.60         & 40.20          \\
IWDD            & \textbf{71.80} & \textbf{44.40} & \textbf{43.40} & {44.60} & \textbf{50.40} & \textbf{43.20} \\ \hline
\end{tabular}}
\label{Tab:auto}
\end{table}

\end{document}